\documentclass[10pt,letterpaper]{article}

\makeatother

\usepackage{comment}

\usepackage{times}
\usepackage{epsfig}
\usepackage{graphicx}
\usepackage{amsmath}
\usepackage{amssymb}

\usepackage{booktabs} 
\usepackage{psfrag}
\usepackage{algorithm}
\usepackage{algorithmic}
\usepackage{subfigure}
\usepackage{url}

\usepackage{lineno}
\usepackage{color}

\begin{document}
\pagestyle{headings}

\newcommand{\nwa}{}

\newcommand{\ie}{\emph{i.e.\@ }}
\newcommand{\eg}{\emph{e.g.\@ }}
\newcommand{\etal}{et al.\@ }
\newcommand{\algoRefText}{Alg.\@ }
\newcommand{\algosRefText}{Algs.\@ }
\newcommand{\figRefText}{Fig.\@ }
\newcommand{\figsRefText}{Figs.\@ }
\newcommand{\eqRefText}{Eq.\@ }
\newcommand{\eqsRefText}{Eqs.\@ }

\newcommand{\siftdist}{$\text{SIFT}_{\text{\tiny{DIST}}}$}
\newcommand{\emdl}{EMD-$L_1$}
\newcommand{\newemdname}{\widehat{\text{EMD}}}

\newcommand{\distfamily}{Quadratic-Chi}
\newcommand{\dd}{QC}
\newcommand{\dda}{QCN}
\newcommand{\ddaname}{Quadratic-Chi-Normalized}
\newcommand{\ddb}{QCS}
\newcommand{\ddbname}{Quadratic-Chi-Squared}
\newcommand{\df}{$\text{QF}$}
\newcommand{\crossbindist}{\mathcal{D}}

\newcommand{\ddns}{NSI-QC}
\newcommand{\ddans}{NSI-QCN}
\newcommand{\ddbns}{NSI-QCS}
\newcommand{\ddnq}{NQI-QC}
\newcommand{\ddanq}{NQI-QCN}
\newcommand{\ddbnq}{NQI-QCS}

\newcommand{\metricpropertiesdist}{\mathcal{D}}

\newcommand{\firsteqssize}{\footnotesize}

\newcommand{\tq}{\texttt{query}}
\newcommand{\ta}{(\texttt{a})}
\newcommand{\tb}{(\texttt{b})}
\newcommand{\tc}{(\texttt{c})}

\newcommand{\histsupperbound}{U}

\newcommand{\dsca}{$\text{\dda}^{1-\frac{\text{dsc}_{T=2}}{2}}$}
\newcommand{\dscb}{$\text{\dda}^{I}$}
\newcommand{\dscc}{$\text{\ddb}^{1-\frac{\text{dsc}_{T=2}}{2}}$}
\newcommand{\dscd}{$\text{\ddb}^{I}$}
\newcommand{\dsce}{$\chi^2$}
\newcommand{\dscf}{$QF^{1-\frac{\text{dsc}_{T=2}}{2}}$}
\newcommand{\dscg}{$L_2$}
\newcommand{\dsch}{$\newemdname{}^{\text{dsc}_{T=2}}_{1}$}
\newcommand{\dsci}{$L_1$}
\newcommand{\dscj}{\siftdist}
\newcommand{\dsck}{\emdl}
\newcommand{\dscl}{Diffusion}
\newcommand{\dscm}{Bhattacharyya}
\newcommand{\dscn}{KL}
\newcommand{\dsco}{JS}

\newcommand{\logand}{\wedge}
\newcommand{\logor}{\vee}
\newcommand{\coldist}{$\text{COL}_{\text{\tiny{DIST}}}$}
\newcommand{\deZeroZero}{\Delta E_{00}}

\newcommand{\distgroupa}{\emph{very-similar}}
\newcommand{\distgroupb}{\emph{similar}}
\newcommand{\distgroupc}{\emph{different}}



\title{Improving Perceptual Color Difference using Basic Color Terms}



\author{Ofir Pele\\
The University of Pennsylvania\\
{\tt\small ofirpele@cis.upenn.edu}
\and
Michael Werman\\
The Hebrew University of Jerusalem\\
{\tt\small werman@cs.huji.ac.il}
}

\maketitle

\begin{abstract}

  We suggest a new color distance based on two observations. First, perceptual color differences
  were designed to be used to compare very similar colors. They do not capture human perception for
  medium and large color differences well. Thresholding was proposed to solve the problem for large
  color differences, \ie two totally different colors are always the same distance apart. We show
  that thresholding alone cannot improve medium color differences. We suggest to alleviate this
  problem using basic color terms. Second, when a color distance is used for edge detection, many
  small distances around the just noticeable difference may account for false edges. We suggest to
  reduce the effect of small distances.

\end{abstract}

\section{Introduction}

\begin{figure}[h!] \centering
\psfrag{distance}[c][c]{{\footnotesize distance}}
\psfrag{distgraphtitle}[c][c]{{\footnotesize Euclidean distance to blue in L*a*b* space}}
\includegraphics[width=0.7\textwidth]{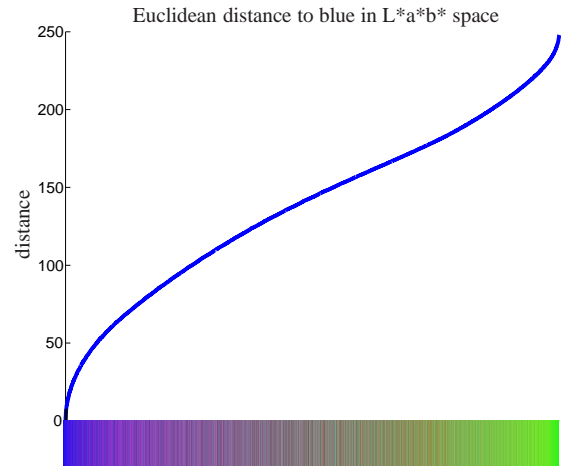}
\caption{ This figure should be viewed in color, preferably on a computer screen. \newline 
The x-axis is colors image, where the colors are sorted by their Euclidean distance in L*a*b* space to the blue color. We can see that distances up to 7 CIELAB units 
capture just a small fraction of similar colors (zoom in to see in the left a black line touching the y-axes in 7 CIELAB units). 
}
\label{euc7}
\end{figure}

Color difference perception of just notably different colors and defining distances between
very-similar colors has received considerable work
\cite{macadam1942vsc,robertson1990hdc,ciede2000,sharma2005ccd,color_book}. In the CIE
(International Commission on Illumination) community, distances of up to 7 CIELAB units, where 1
CIELAB unit approximately correspond to 1 just noticeable difference, are considered medium
distances \cite{large_color_diffs}. In this paper we refer to 0-7 CIELAB distances as very-similar
as they capture just a small fraction of similar colors. 
See \figRefText \ref{euc7}.

The CIEDE2000 color difference is considered the state of the art perceptual color difference
\cite{ciede2000,ciede2000_test_on_crt,large_color_diffs}. CIEDE2000's recommended range for use is
0 to 5 CIELAB units \cite{cie_recommendation_5_units}. The COM dataset was used to train and
perceptually test the CIEDE2000 color difference. More than 95 percent of the distances between
color pairs in the COM dataset are below 5 CIELAB units apart.
 
It was pointed out that the resulting color differences do not correspond well with human
perception for medium to large distances. Rubner \etal \cite{rubner_emd} and Ruzon and Tomasi
\cite{ruzon_compass} used a negative exponent on the color difference. Namely, all totally
different colors are essentially assigned the same large distance. Pele and Werman
\cite{Pele-iccv2009} noted that a negative exponent changes the values in the small distance
range. Pele and Werman \cite{Pele-iccv2009} and Rubner \etal \cite{rubner_emd} observed a reduction
in performance due to this change. Pele and Werman suggested thresholding the color difference as
it does not change the small distances. Thresholding color distances is justified by the fact that
if people are directly asked for a judgment of the dissimilarity of colors far apart in color
space, subjects typically find themselves unable to express a more precise answer than ``totally
different'' \cite{indow1994metrics}.
An additional advantage of thresholding color distances is that it allows fast computation of cross-bin distances such as the Earth Mover's Distance \cite{Pele-iccv2009} or using the transformation to a similarity measure of one minus the distance divided by the threshold, the Quadratic-Chi \cite{peleeccv2010}.


\begin{figure}[h!] \centering
\psfrag{similar}[c][c]{{\footnotesize similar}}
\psfrag{different}[c][c]{{\footnotesize different}}
\psfrag{colors}[c][c]{{\footnotesize colors}}
\psfrag{(a)}[c][c]{(a)}
\psfrag{(b)}[c][c]{(b)}
\psfrag{(c)}[c][c]{(c)}
\psfrag{(d)}[c][c]{(d)}
\psfrag{(e)}[c][c]{(e)}
\psfrag{(f)}[c][c]{(f)}
\includegraphics[width=0.8\textwidth]{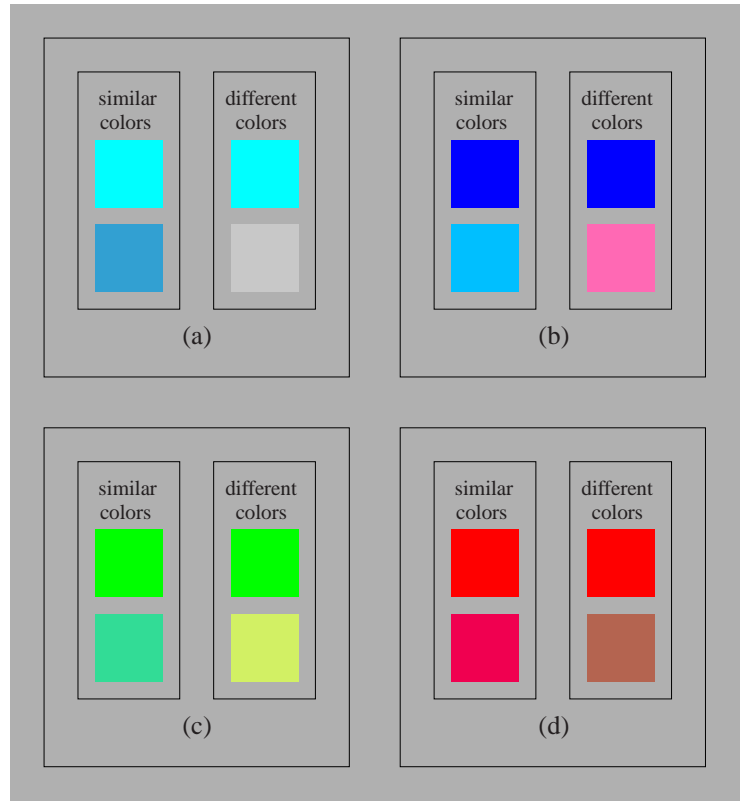}
\caption{ This figure should be viewed in color, preferably on a computer screen. \newline Each
  sub-figure (a)-(d) contains a pair of similar colors and a pair of different colors. 
  In all of these examples, the CIEDE2000 distance between the visually similar
  colors is higher than the distance between the different colors. 
  Our proposed distance succeeds in all of these examples.
}
\label{patchesFig}
\end{figure}

This paper shows that CIEDE2000 is not a good distance for the medium range and using any monotonic
function of CIEDE2000 (including a thresholding function) cannot solve the problem. For example, using a
thresholding function we cannot make DarkSkyBlue be more similar to Blue than to HotPink. See
\figRefText \ref{patchesFig} for more examples.  

We suggest an improvement based on basic color terms. Specifically we use Berlin and Kay's eleven
English basic color terms\cite{berlin_kay}. However, the generalization to other color terms is
straight forward. We suggest adding to color differences the distance between their basic
color terms probability vectors. As basic color terms are correlated (\eg red and orange) we
suggest using a cross-bin distance for these probability vectors. That is, a distance which takes
the relationships between bins (each bin represents a basic color term) into account. Specifically
we use the Earth Mover's Distance \cite{rubner_emd} as it was used successfully in many
applications (\eg \cite{rubner_emd,rubner_emd_comparison,ruzon_compass,Pele-iccv2009} and
references within).

The probability vectors are obtained with the color naming method developed by van de Weijer \etal
\cite{vandeweijer2009lcn}. Other methods for color naming such as
\cite{conway1992experimental,lammens1994computational,seaborn1999fuzzy,benaventeBOV00,griffin2004optimality,mojsilovic2005computational,benavente2006data,menegaz2006discrete,menegaz2007semantics,benavente2008parametric}
can also be used. We chose van de Weijer \etal method as it has excellent performance on real-world
images and as the code for it is publicly available. However, CIEDE2000 was learned under
calibrated conditions, while van de Weijer \etal method was learned from natural images. Thus,
other color naming methods might produce better results. This is left for future work.

Our proposed solution is not equivalent to increasing the weight of the hue component in the color
difference. Color names are not equivalent to hue. For example, although a rainbow spans a
continuous spectrum of colors, people see in it distinct bands which correspond to basic color
terms: red, orange, yellow, green, blue and purple. In addition, some basic color terms are not
different in their hue component. \eg achromatic colors such as white, gray and black or orange and
brown which shares the same hue.

A second problem that occurs when using color differences for edge detection is that many small
distances around the just noticeable difference may account for false edges. We suggest to use a
sigmoid function to reduce the small distances effect. As we mentioned before, using a negative
exponent function in order to assign to all totally different color pairs the same distance reduced
performance \cite{rubner_emd,Pele-iccv2009}. We explain this by the fact that a negative exponent
is a concave function. We show that a convex function should be applied to small differences.

We present experimental results for color edge detection. We show that by using our new color
difference the results are perceptually more meaningful.

Our solution is just the first step of designing a perceptual color difference for the full range of
distances. Our major contribution is raising the problem of current state-of-the-art color
differences in the small and medium distance range.

This paper is organized as follows. Section \ref{related_work} is an overview of related
work. Section \ref{colDist_sec} introduces the new color difference. Section \ref{results_sec}
presents the results. Finally, conclusions are drawn in Section \ref{conclusions_sec}.

\begin{figure*}[t] \centering
\psfrag{a}[r][c]{{\tiny \coldist}}
\psfrag{b}[r][c]{{\tiny CIEDE2000}}
\includegraphics[width=0.95\textwidth]{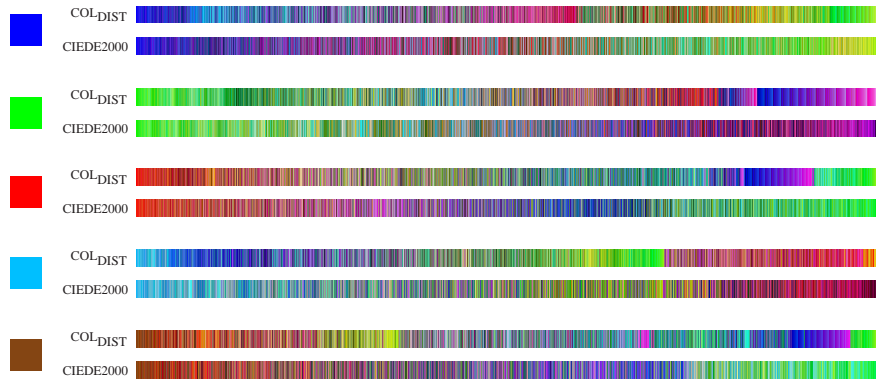}
\caption{ This figure should be viewed in color, preferably on a computer screen. Use the pdf
  viewer's zoom to see the colors.  We show colors, sorted by their distance to the color on the
  left. \coldist{} is our new perceptual color difference. Several observations can be derived from
  these graphs. First, our distance is perceptually better in the medium distance range. Note that
  the group of similar colors (left side of each color legend) is more similar to the color on the
  left using our distance. For example, in the top light blues are similar to blue, while using CIEDE2000
  they are very different (thus appear on the right). It should be noted that our
  distance uses a sigmoid function, so that very similar colors on the left are essentially
  assigned the same small distance and totally different colors on the right are essentially
  assigned the same large distance. Finally, although our distance is perceptually more
  meaningful, it is still far from being perfect.  }
\label{color_dist_graph}
\end{figure*}

\section{Related Work}
\label{related_work}

MacAdam's \cite{macadam1942vsc} pioneering work on chromaticity discrimination ellipses, which
measured human perception of just noticeable differences led the way to the development of the
L*a*b* space \cite{robertson1990hdc} which is considered perceptually uniform; \ie for very-similar
colors, the Euclidean distance in the L*a*b* space corresponds to the human perception of color
difference well. Luo \etal \cite{ciede2000} developed the CIEDE2000 color difference which is now
considered the state of the art perceptual color difference
\cite{ciede2000,ciede2000_test_on_crt,large_color_diffs}.

Although color is commonly experienced as an indispensable quality in describing the world around
us, state-of-the art computer vision methods are mostly based on shape description and ignore
color information. Recently this has changed with the introduction of new color descriptors
\cite{color_features_1,vdw12,vdw13,vdw11,vandeweijer2006clf,BurghoutsCVIU2009,songlocal}. However, although color is a
point-wise property (\eg bananas are yellow), most of these features capture geometric relations
such as color edges.

Wertheimer \cite{wert} suggested that among perceptual stimuli there are ``ideal types'' that are
anchor points for perception. Rosch \cite{rosch1975cognitive} proposed that in certain perceptual
domains, such as color, salient prototypes develop non arbitrarily. An influential paper by Berlin and
Kay \cite{berlin_kay} defined basic colors as color names in a language which are applied to
diverse classes of objects and whose meaning is not subsumable under one of the other basic color
names and which are used consistently and by consensus by most of the speakers of the language. In
their pioneering anthropological study, they found that color was usually partitioned into a
maximum of eleven basic color categories of which three were achromatic (black, white, grey) and
eight chromatic (red, green, yellow, blue, purple, orange, pink and brown). This partitioning was a
universal tendency to group color around specific focal points as was conjectured by Wertheimer
\cite{wert} and Rosch \cite{rosch1975cognitive}.


Considerable work has been carried out in the field of computational color naming, see \eg
\cite{conway1992experimental,lammens1994computational,seaborn1999fuzzy,benaventeBOV00,griffin2004optimality,mojsilovic2005computational,benavente2006data,menegaz2006discrete,menegaz2007semantics,benavente2008parametric,vandeweijer2009lcn}
and references within. Recently van de Weijer \etal \cite{vandeweijer2009lcn} presented a new color
naming method based on real-world images. The color names are Berlin and Kay's \cite{berlin_kay}
eleven English basic color terms. Van de Weijer and Schmid \cite{vandeweijer2007acn} showed that a
color description based on these color names outperforms descriptions based on photometric
invariants. The explanation is that photometric invariance reduces the discriminative power of the
descriptor.

Inspired by van de Weijer and Schmid's work we suggest using the basic color names to correct the
state-of-the-art color difference, CIEDE2000 in the medium distance range.

\section{\coldist : The New Color Difference}
\label{colDist_sec}

Given two colors $C^1=[R^1,G^1,B^1]$ and $C^2=[R^2,G^2,B^2]$, we first convert\footnote{We used
  Matlab's default conversion which uses CIE illuminant D50 as the reference illuminant, known as
  the white point, which is also the default illuminant specified in the International Color
  Consortium specifications.} them into L*a*b* : $S^1=[L^1,a^1,b^1]$ and
$S^2=[L^2,a^2,b^2]$. Second, we compute the basic color term probability vectors: $P^1,P^2$; where
$P^{n}_i$ is the probability that the color $C^n$ is the basic color term $i$ (\ie black, blue,
brown, grey, green, orange, pink, purple, red, white or yellow). These probability vectors are
computed using the van de Weijer \etal color naming method \cite{vandeweijer2009lcn}. Now each
color $C^n$ is represented by an 14-dimensional vector: $V^n=[S^n,P^n]=[L^n , a^n , b^n, P^n_{1},
\ldots,P^n_{11}]$.  The distance between the two colors (parameterized with $T$, $D$, $\alpha$ and
$Z$) is defined as:

\begin{align}
d_1(S^1,S^2)&= \frac{\min( \text{CIEDE2000}(S^1,S^2) , T )}{T} \label{d1_eq} \\
d_2(P^1,P^2)&= \text{EMD}(P^1,P^2,D) \label{d2_eq} \\
d_3(V^1,V^2)&= \alpha d_1 + (1-\alpha)d_2 \label{d3_eq} \\
\text{\coldist}& (V^1,V^2)=  \frac{1}{1+e^{-(Zd_3 - \frac{Z}{2})}} \label{coldist_eq}
\end{align}


In \eqRefText \ref{d1_eq}, $d_1$ is a thresholded and scaled CIEDE2000 color difference. We threshold
it as it is recommended for use only for small distances \cite{cie_recommendation_5_units}. We
used $T=20$ as was used in Pele and Werman \cite{Pele-iccv2009}. We divide by $T$ so that $d_1$
is between 0 and 1.

In \eqRefText \ref{d2_eq} $d_2$ is the distance between the two basic colors probability vectors.
As the bins in the eleven basic colors probability vectors are correlated (\eg orange and red), we
use the Earth Mover's Distance that takes this correlation into account. The correlation is encoded
in $D$ which is an $11 \times 11$ matrix, where $D_{ij}$ is the distance between basic color term
$i$ to basic color term $j$. We estimated $D$ using the joint distribution of the basic color
terms. That is, given the matrix $M$ of all probability vectors for the colors in the RGB cube
($2^{15} \times 11$ matrix, as each dimension of the RGB cube was quantized with jumps of $8$
\cite{vandeweijer2009lcn}) we define $D=D_{ij}$ as:

\begin{align}
\hat{D}_{ij}&= 1 - 2 \left( \frac{\sum_{n} \min( M_{ni},M_{nj} ) }{ \sum_{n} M_{ni}+M_{nj} } \right) \\
D_{ij}&= \frac{ \min (\hat{D}_{ij},t) }{t}
\end{align}

\begin{figure}[h!] \centering
\psfrag{0}{\scriptsize 0}
\psfrag{0.1}{\scriptsize 0.1}
\psfrag{0.2}{\scriptsize 0.2}
\psfrag{0.3}{\scriptsize 0.3}
\psfrag{0.4}{\scriptsize 0.4}
\psfrag{0.5}{\scriptsize 0.5}
\psfrag{0.6}{\scriptsize 0.6}
\psfrag{0.7}{\scriptsize 0.7}
\psfrag{0.8}{\scriptsize 0.8}
\psfrag{0.9}{\scriptsize 0.9}
\psfrag{1}{\scriptsize 1}
\psfrag{black}[c][l]{\scriptsize black}
\psfrag{blue}[c][l]{\scriptsize blue}
\psfrag{brown}[c][l]{\scriptsize brown}
\psfrag{grey}[c][l]{\scriptsize grey}
\psfrag{green}[c][l]{\scriptsize green}
\psfrag{orange}[c][l]{\scriptsize orange}
\psfrag{pink}[c][l]{\scriptsize pink}
\psfrag{purple}[c][l]{\scriptsize purple}
\psfrag{red}[c][l]{\scriptsize red}
\psfrag{white}[c][l]{\scriptsize white}
\psfrag{yellow}[c][l]{\scriptsize yellow}
\psfrag{blackx}[c][c]{\scriptsize black}
\psfrag{bluex}[c][c]{\scriptsize blue}
\psfrag{brownx}[c][c]{\scriptsize brown}
\psfrag{greyx}[c][c]{\scriptsize grey}
\psfrag{greenx}[c][c]{\scriptsize green}
\psfrag{orangex}[c][c]{\scriptsize orange}
\psfrag{pinkx}[c][c]{\scriptsize pink}
\psfrag{purplex}[c][c]{\scriptsize purple}
\psfrag{redx}[c][c]{\scriptsize red}
\psfrag{whitex}[c][c]{\scriptsize white}
\psfrag{yellowx}[c][c]{\scriptsize yellow}
\includegraphics[width=0.95\textwidth]{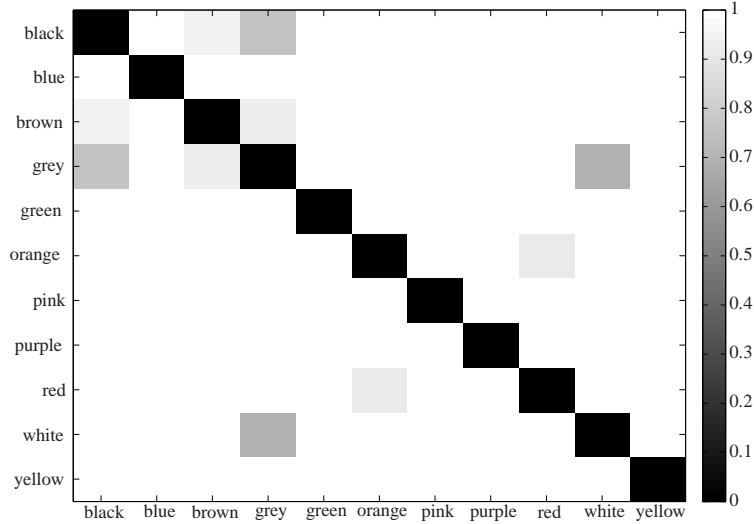}
\caption{ 
The learned basic color terms ground distance matrix.
}
\label{basic_color_ground_distance_fig}
\end{figure}

We threshold $\hat{D}_{ij}$ as EMD is recommended to be used with thresholded ground distances
\cite{Pele-iccv2009}. We used the threshold $t=0.7$. Finally we scale it so that $0 \leq D_{ij}
\leq 1$ (which implies $0 \leq d_2 \leq 1$). The resulting matrix (see \figRefText
\ref{basic_color_ground_distance_fig}) is perceptually plausible; \ie similar basic color terms
are: grey and white, grey and black, orange and red, etc.

The Earth Mover's Distance (EMD) \cite{rubner_emd} is defined as the minimal cost that must be paid
to transform one histogram into another, where there is a ``ground distance'' (that is, the matrix
$D$) between the basic features that are aggregated into the histogram. Here the basic features are
the eleven English basic color terms. The formula for the EMD between the two probability vectors
$P^1$ and $P^2$ is defined as\footnote{This is a simplification of the original definition for the
  case where the histograms are probability vectors.}:

\begin{align}
\begin{split}
\text{EMD}(P^1,P^2,D)&= \min_{\{F_{ij}\}} \sum_{i,j} F_{ij} D_{ij} \;\;\;\; s.t \\
\sum_j F_{ij} = P^1_i \;\; &, \;\;
\sum_i F_{ij} = P^2_j \;\; , \\
\sum_{i,j} F_{ij} = 1 \;\; &, \;\; F_{ij} \geq 0
\end{split}
\label{EMD_orig_eq}
\end{align}

In \eqRefText \ref{d3_eq}, $d_3$ is a linear combination of $d_1$ and $d_2$. We used
$\alpha=\frac{1}{2}$.

Finally, in \eqRefText \ref{coldist_eq}, the distance is scaled so that it is in the range
$[-\frac{Z}{2},\frac{Z}{2}]$ (we used $Z=10$) and then the logistic function (a sigmoid function)
is finally applied. The sigmoid function reduces the effect of small distances and essentially
gives all totally different colors the same distance.

\section{Results}
\label{results_sec}

In this section we present color edge detection results. We used Ruzon and Tomasi generalized
compass edge detector \cite{ruzon_compass}. We used this method for two reasons. First, the code is
publicly available. Second, the code uses only color cues for the edge detection which enables us to
isolate color difference performance.

Ruzon and Tomasi's method \cite{ruzon_compass} divides a circular window around each pixels in half
with a line segment. Then it computes a sparse color histogram (coined signature in their paper)
for each half and computes the Earth Mover's Distance (EMD) \cite{rubner_emd} between the two
histograms. The EMD uses a ground distance matrix $D$ between the colors. Ruzon and Tomasi
converted the images to L*a*b* and then used a negative exponent of the Euclidean distance as the
ground distance between colors:

\begin{align}
d_e(S^1,S^2)&= 1-e^{ \frac{-||[L^1,a^1,b^1]-[L^2,a^2,b^2]||_2}{\gamma}}
\end{align}

\begin{figure*}[htbp] \centering
\begin{tabular}{cc}
\includegraphics[width=0.3\textwidth]{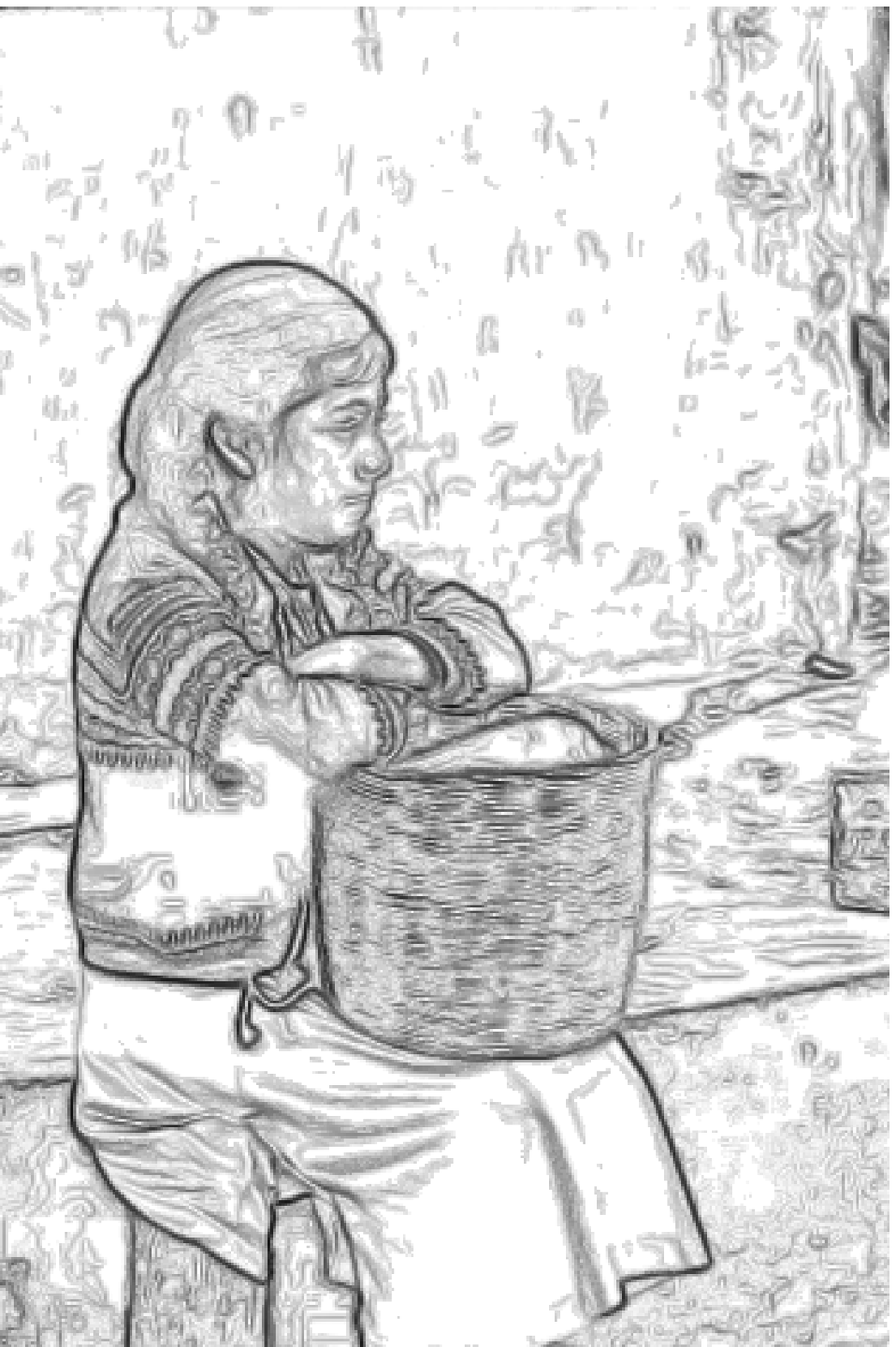} &
\includegraphics[width=0.3\textwidth]{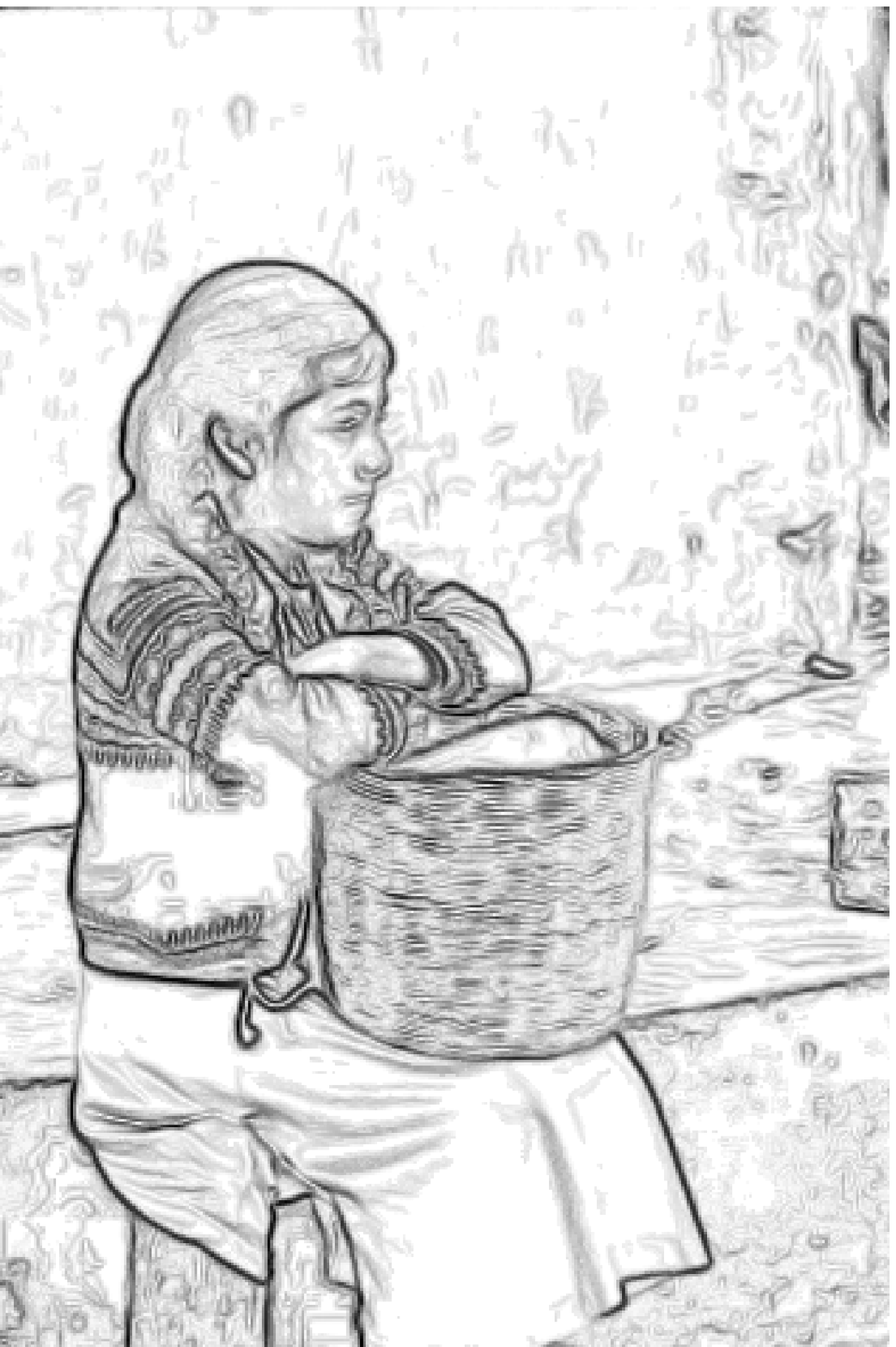} \\
(NE) & (TC) \\
\includegraphics[width=0.3\textwidth]{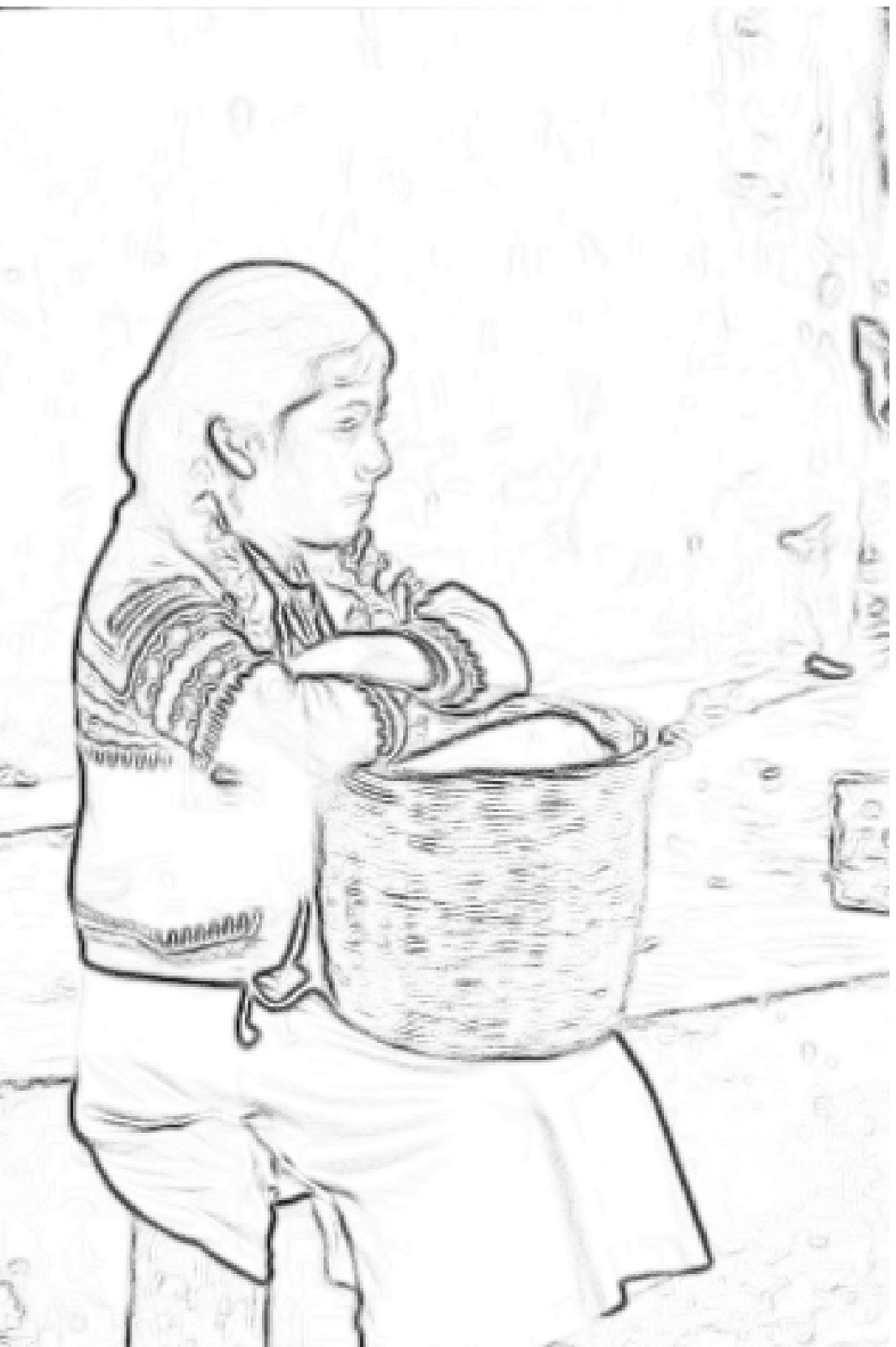} &
\includegraphics[width=0.3\textwidth]{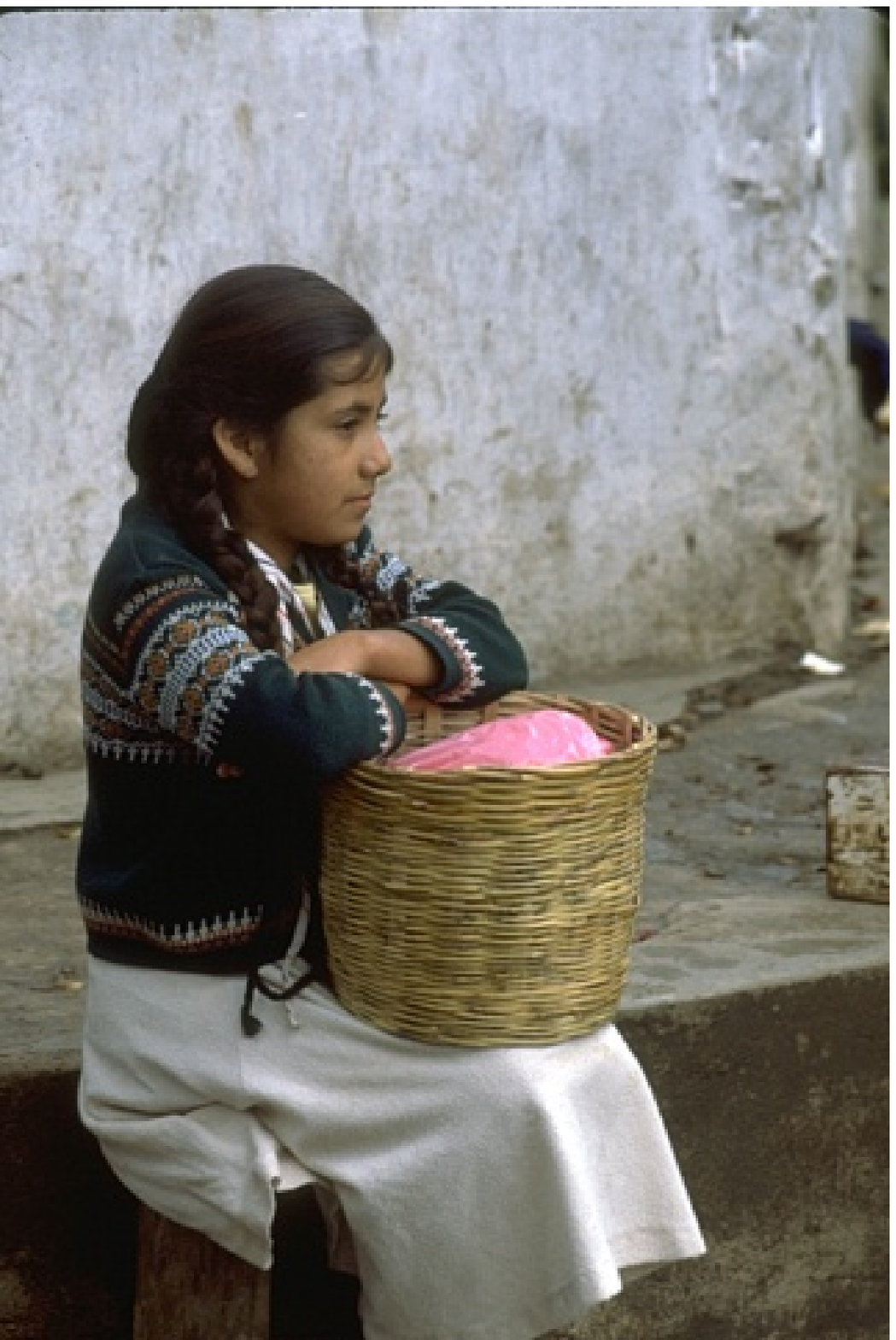} \\
(\coldist) & (IM)
\end{tabular}
\caption{ Edge detection with the generalized compass edge detection \cite{ruzon_compass} using the
  following color differences: (NE) A negative exponent applied on the Euclidean distance in L*a*b*
  space (used in \cite{ruzon_compass}).  (TC) A thresholded CIEDE2000 distance (used in
  \cite{Pele-iccv2009} for image retrieval). See \eqRefText \ref{d1_eq}.  (\coldist) Our proposed
  \coldist.  (IM) The original image. \newline Our result is much cleaner. Note that our method
  detects the right boundary of the basket without detecting many false edges, while in (NE) and
  (TC) the false edges magnitude is larger than the right boundary of the basket. }
\label{res_1}
\end{figure*}

\begin{figure*}[htbp] \centering
\begin{tabular}{cc}
\includegraphics[width=0.45\textwidth]{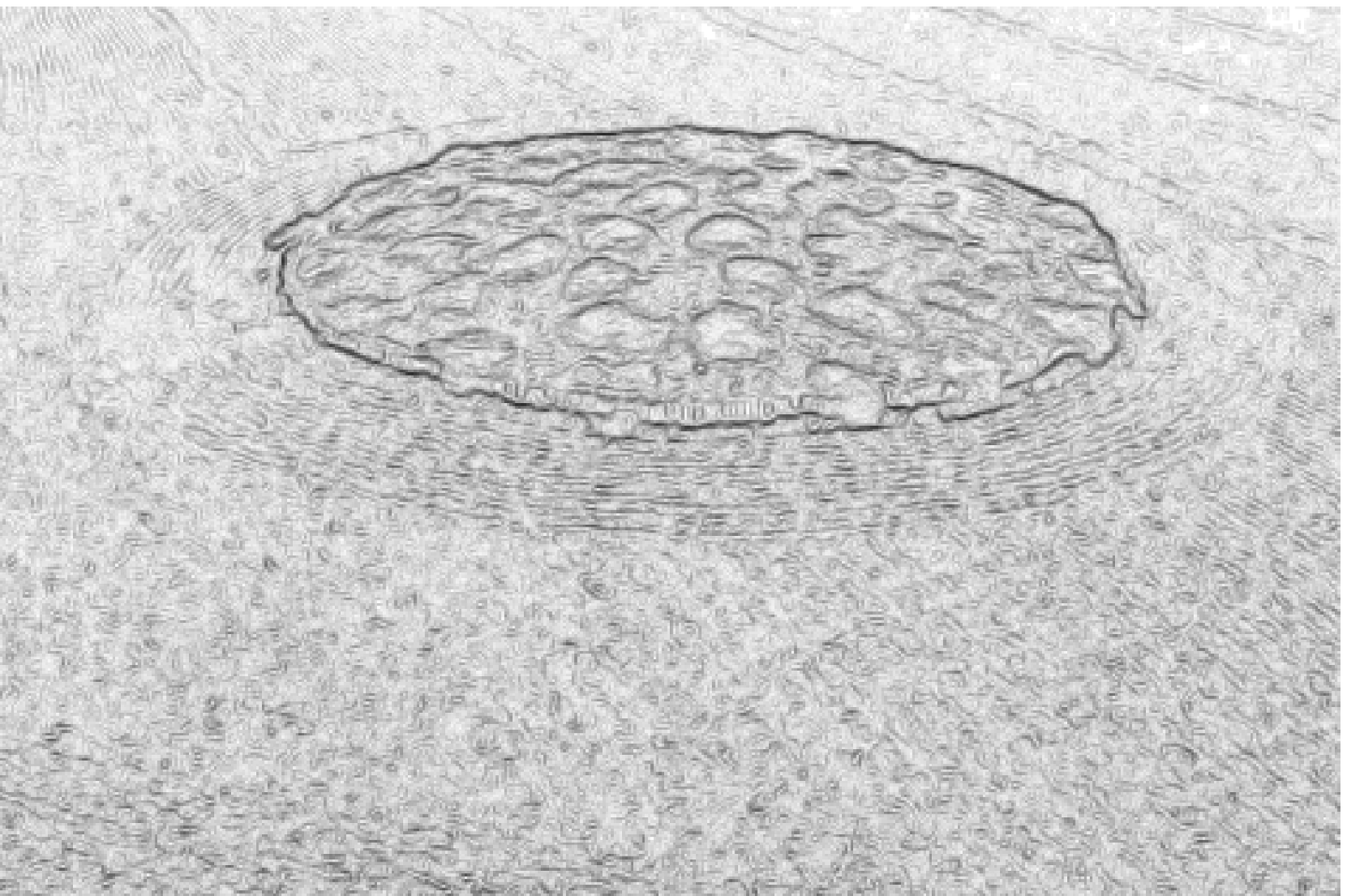} &
\includegraphics[width=0.45\textwidth]{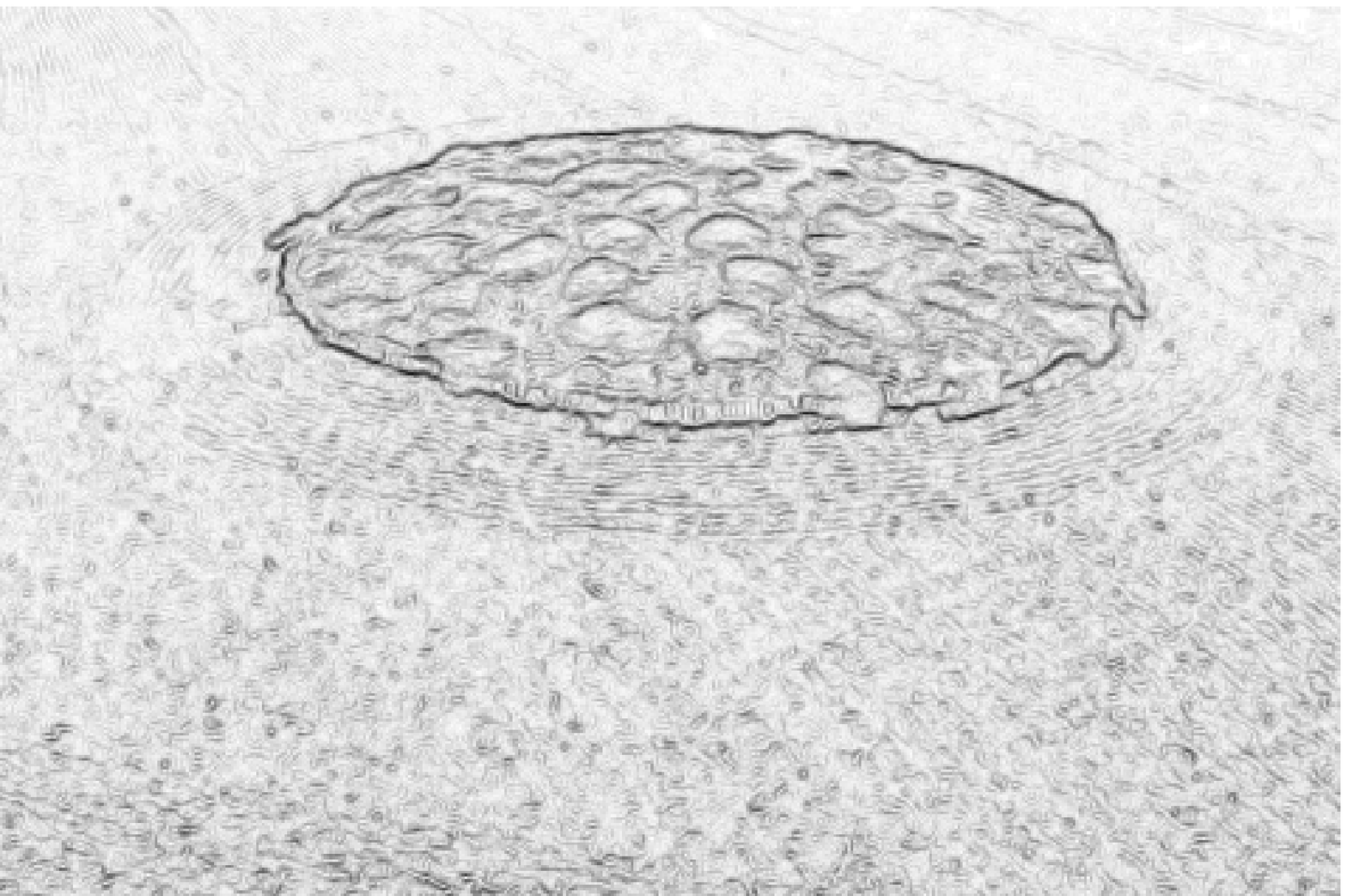} \\
(NE) & (TC) \\
\includegraphics[width=0.45\textwidth]{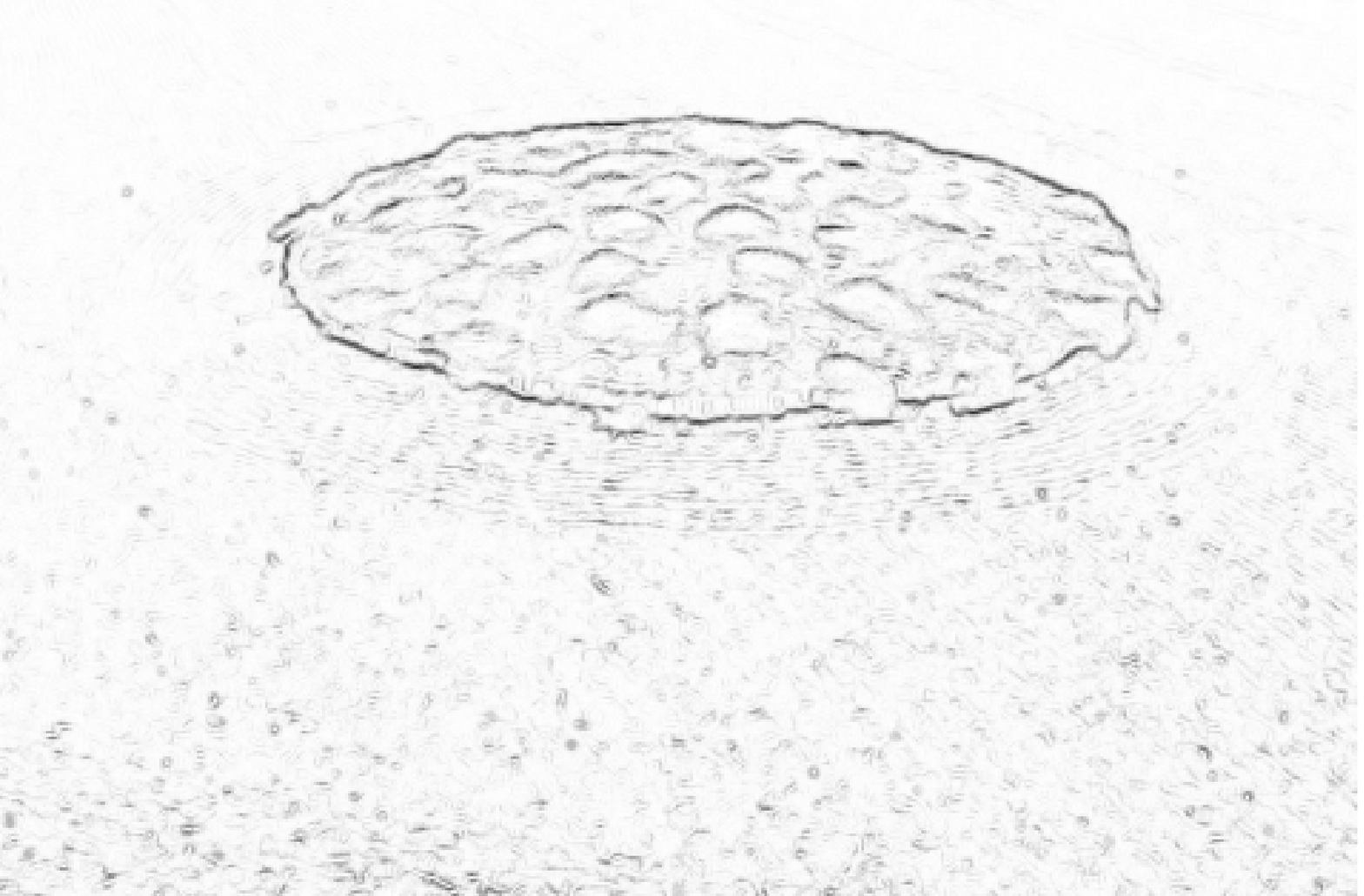} &
\includegraphics[width=0.45\textwidth]{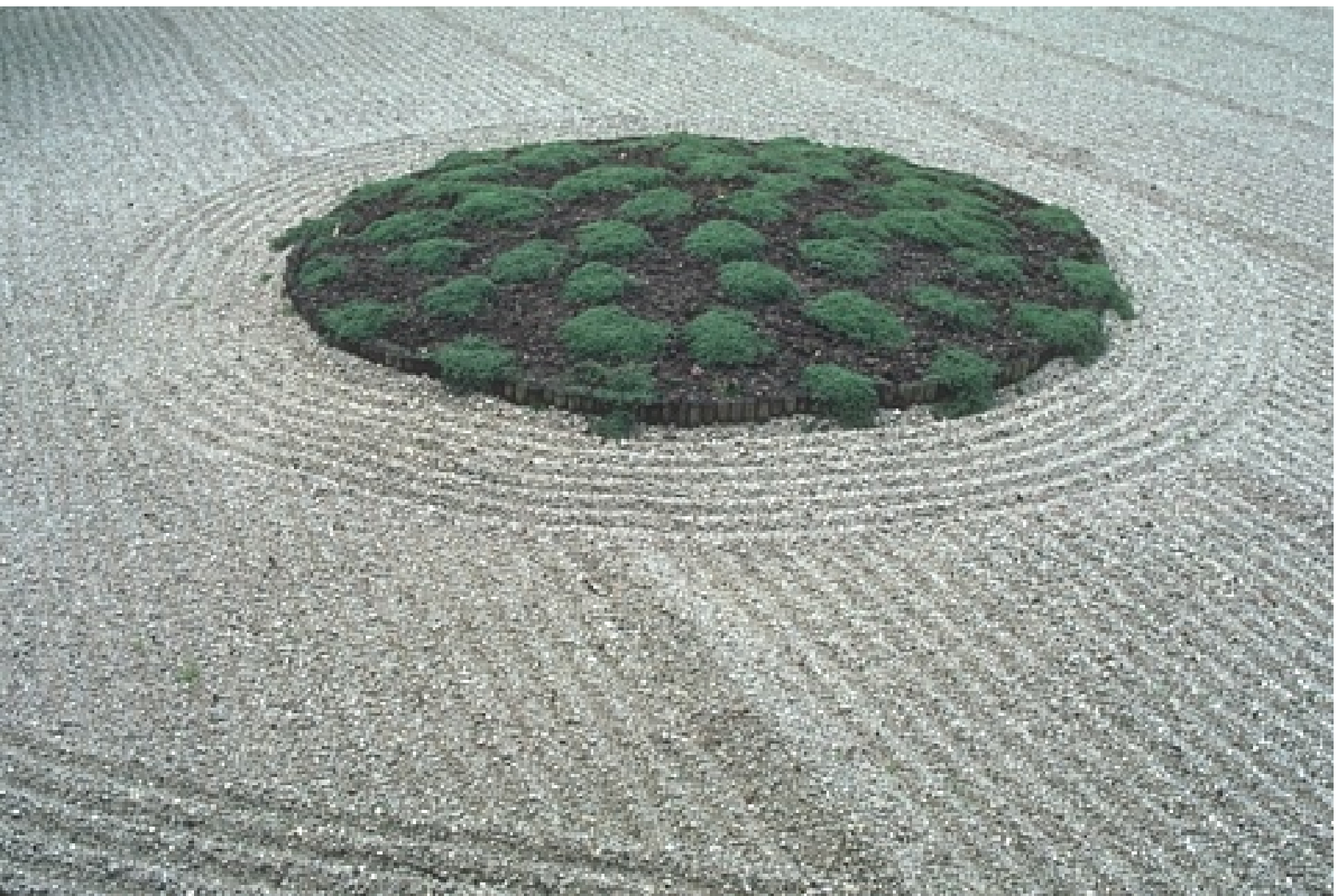} \\
(\coldist) & (IM) \\
\newline \\
\includegraphics[width=0.45\textwidth]{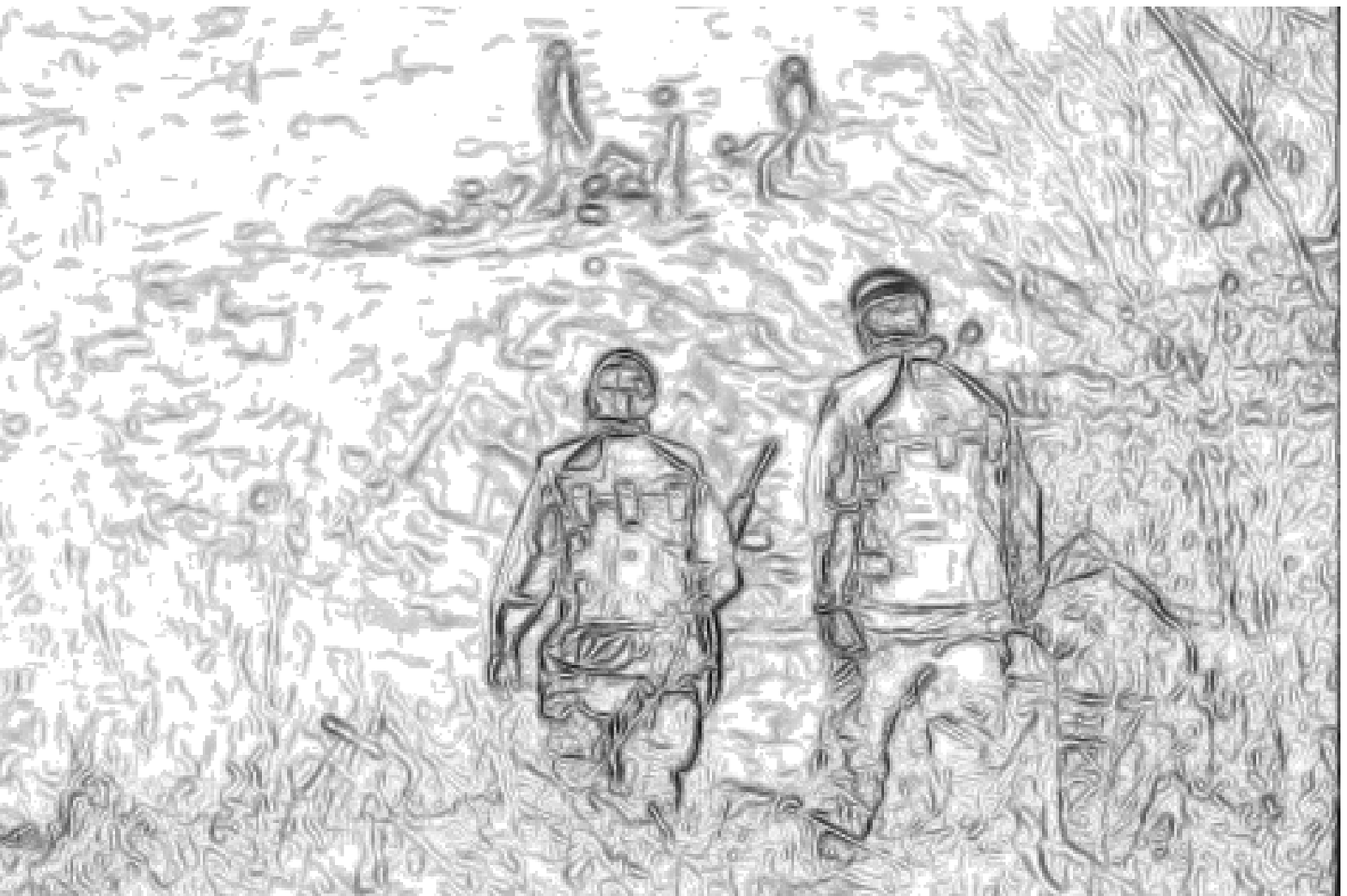} &
\includegraphics[width=0.45\textwidth]{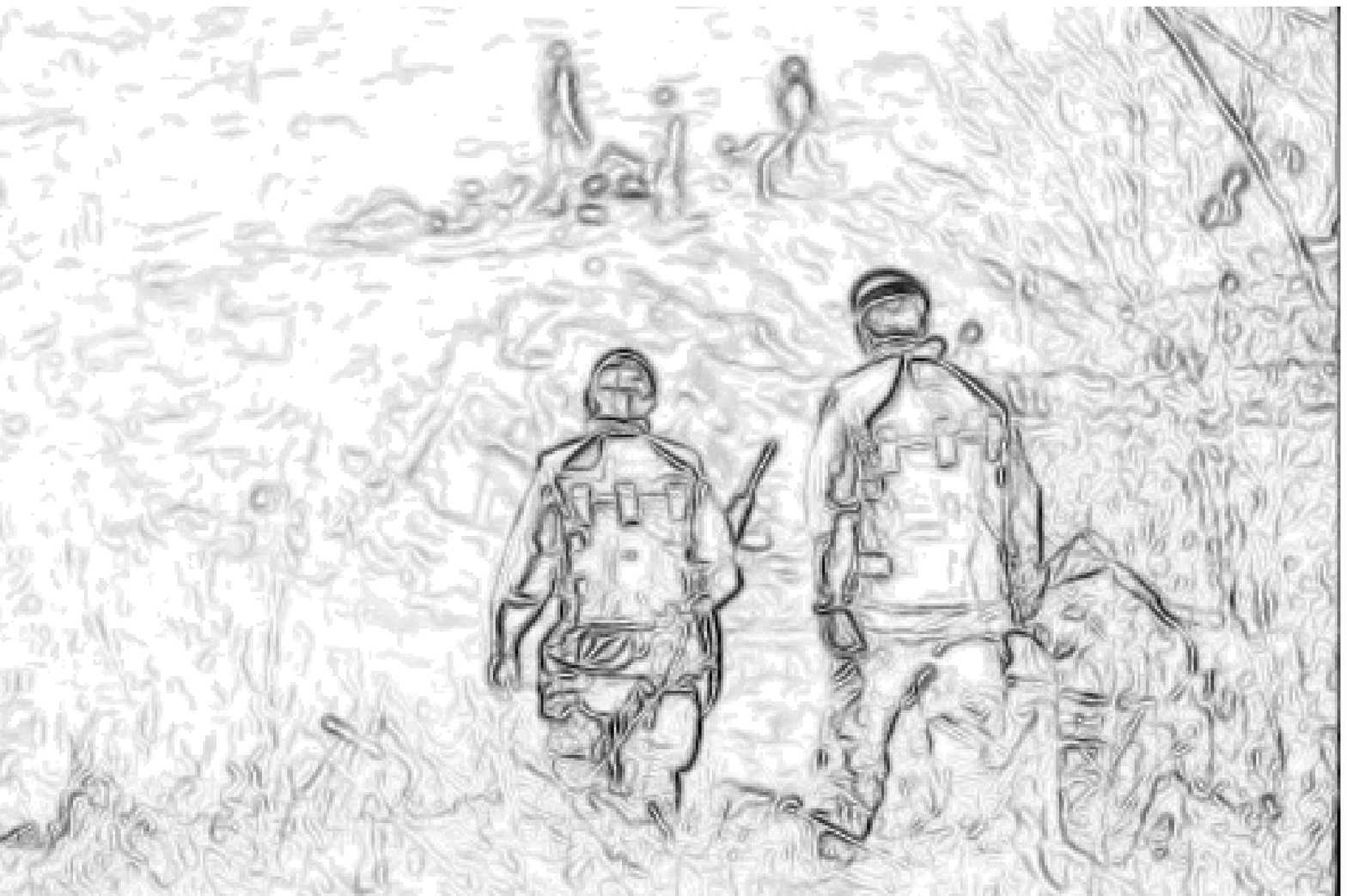} \\
(NE) & (TC) \\
\includegraphics[width=0.45\textwidth]{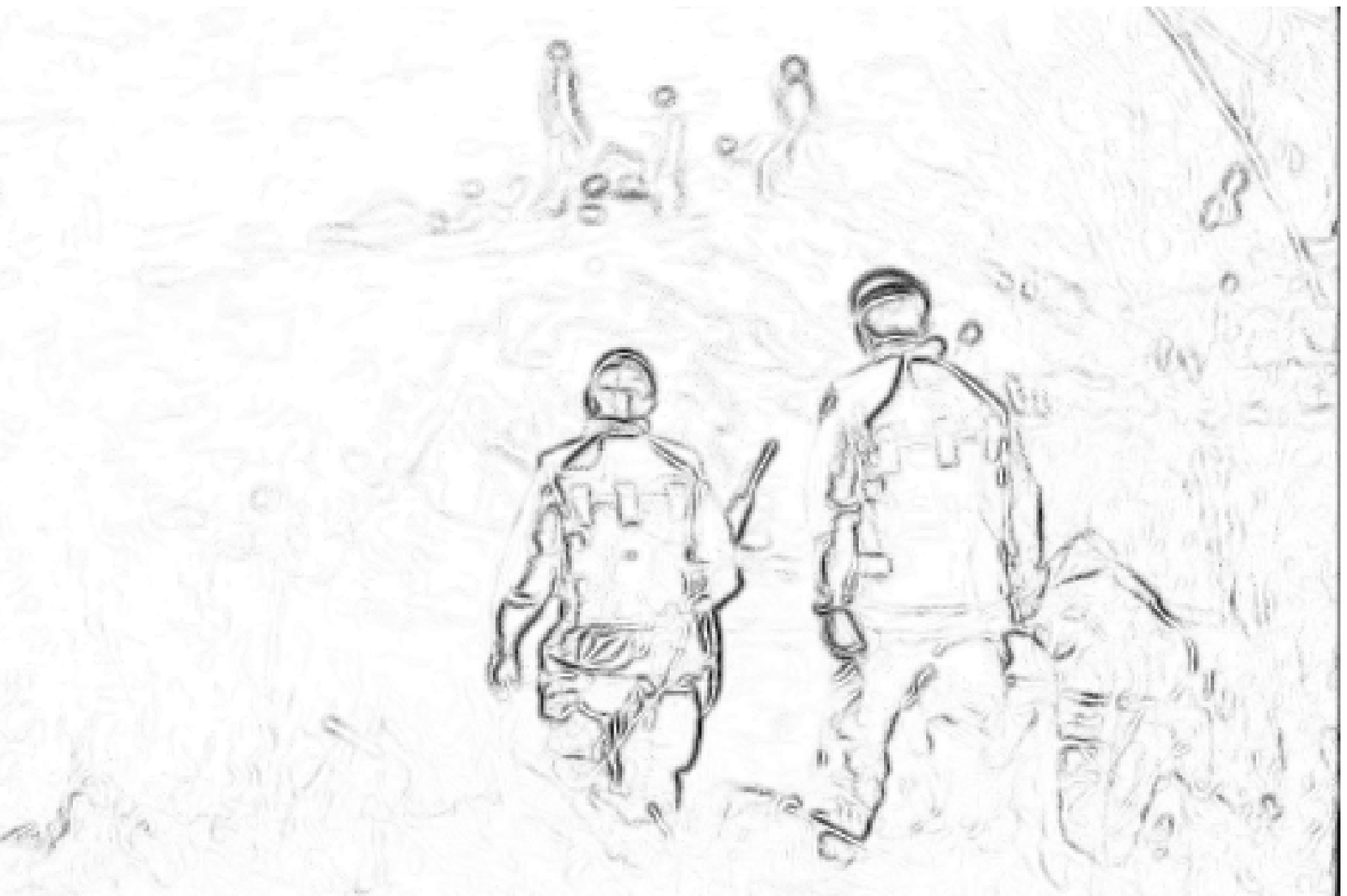} &
\includegraphics[width=0.45\textwidth]{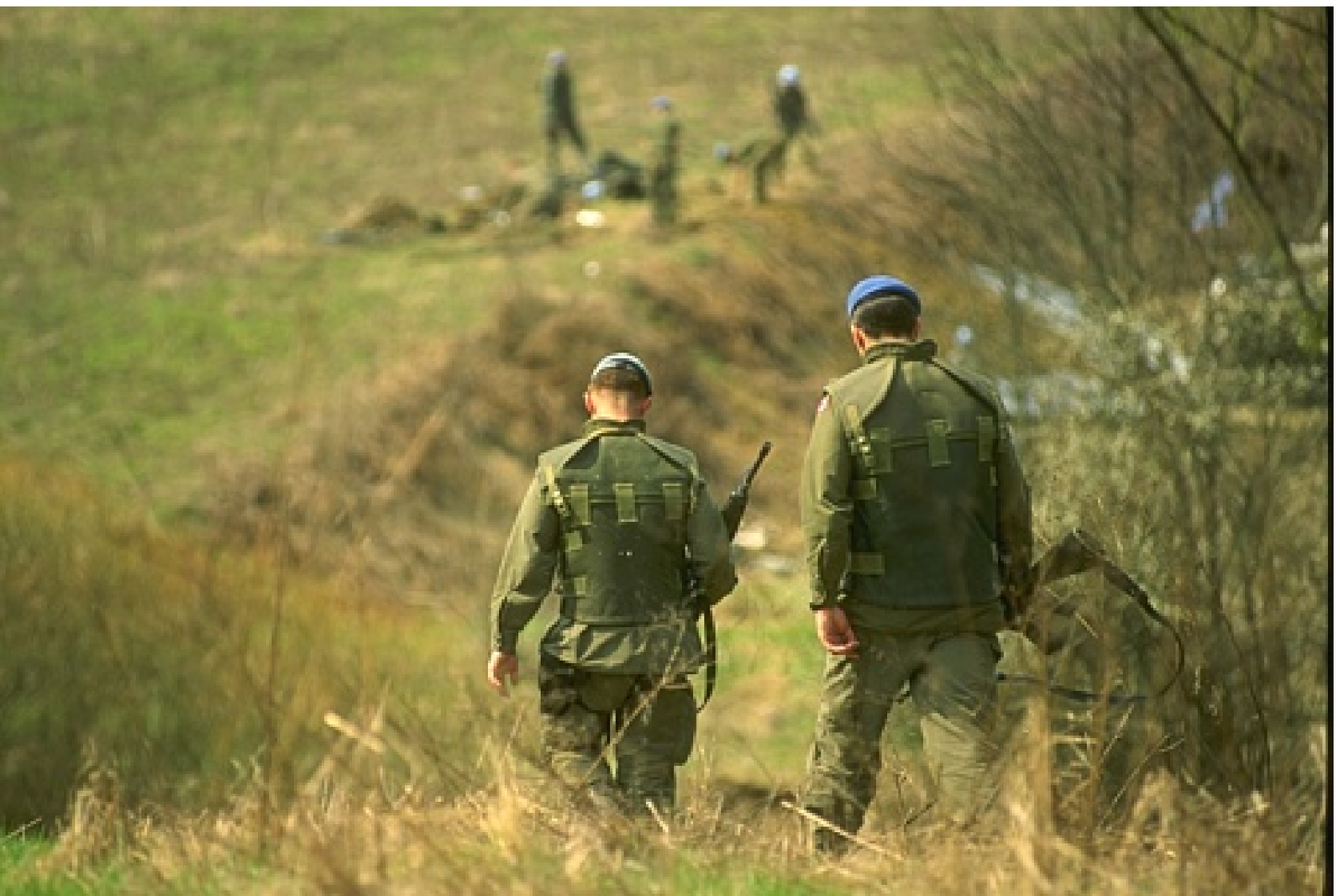} \\
(\coldist) & (IM)
\end{tabular}
\caption{ Edge detection with the generalized compass edge detection \cite{ruzon_compass} using the
  following color differences: (NE) A negative exponent applied on the Euclidean distance in L*a*b*
  space (used in \cite{ruzon_compass}).  (TC) A thresholded CIEDE2000 distance (used in
  \cite{Pele-iccv2009} for image retrieval). See \eqRefText \ref{d1_eq}.  (\coldist) Our
  proposed \coldist.  (IM) The original image. \newline
Our results are much cleaner. Note that on the top the clean detection of the bushes boundaries.
 }
\label{res_2}
\end{figure*}

\begin{figure*}[htbp] \centering
\begin{tabular}{cc}
\includegraphics[width=0.3\textwidth]{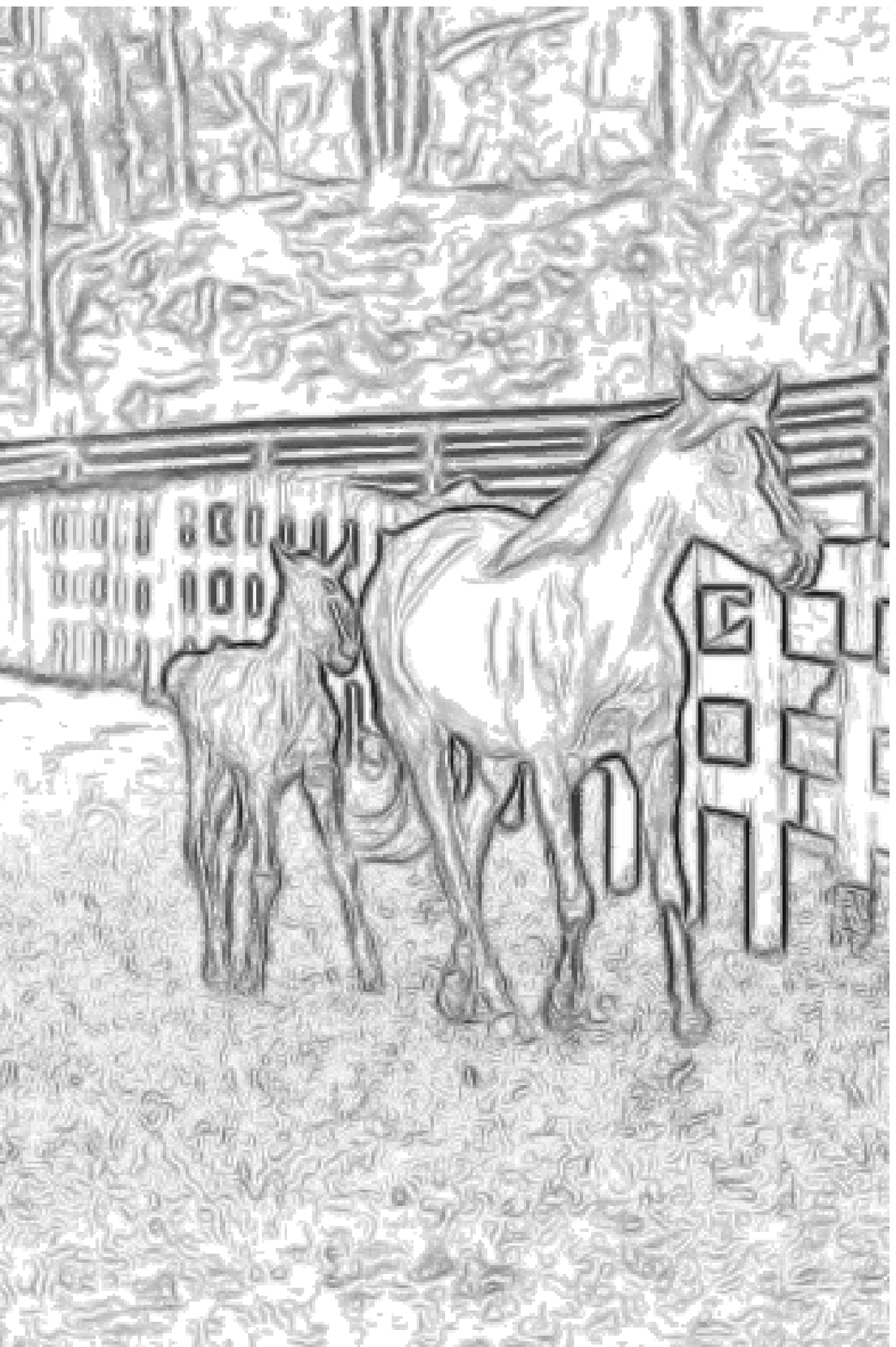} &
\includegraphics[width=0.3\textwidth]{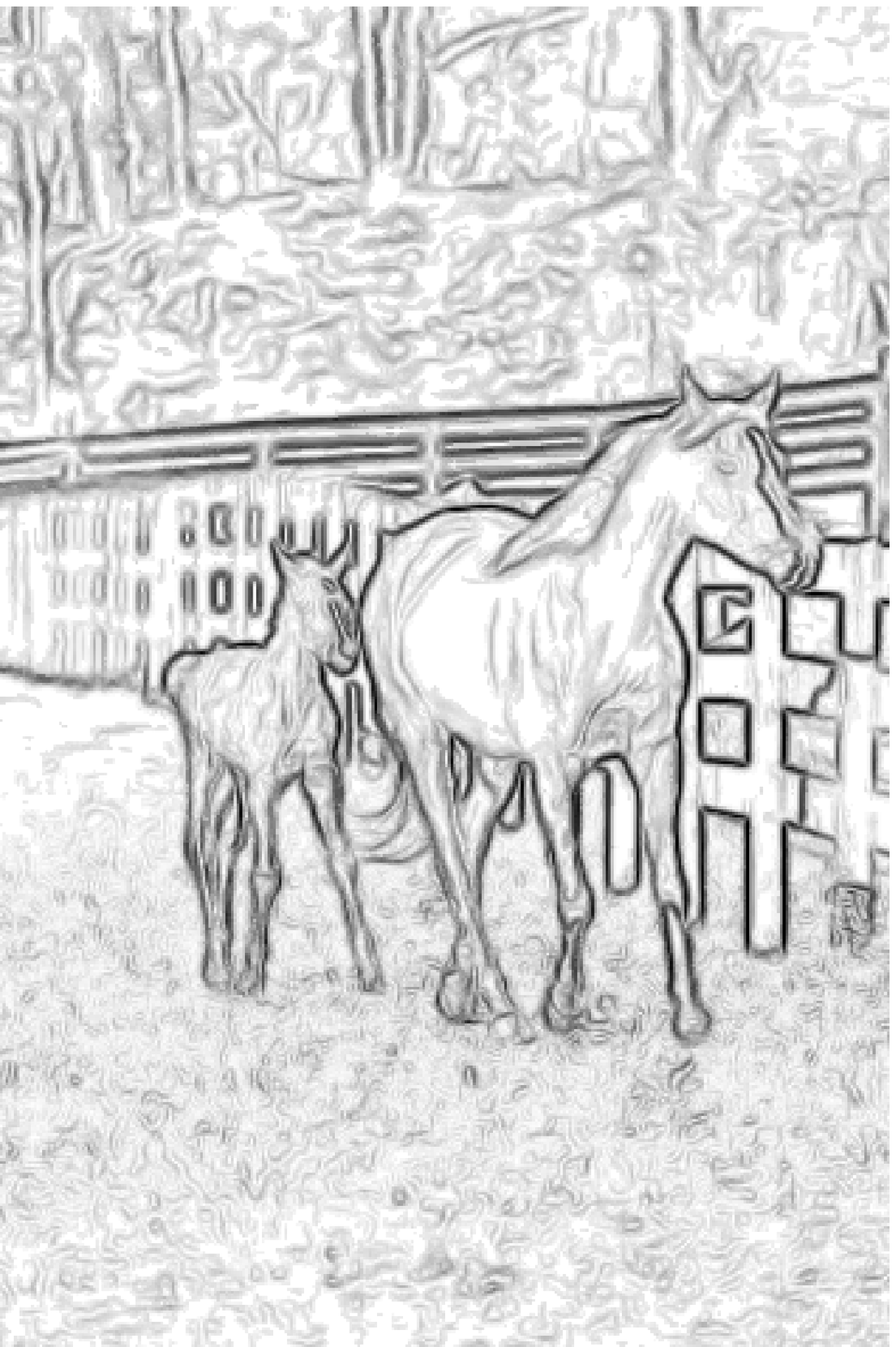} \\
(NE) & (TC) \\
\includegraphics[width=0.3\textwidth]{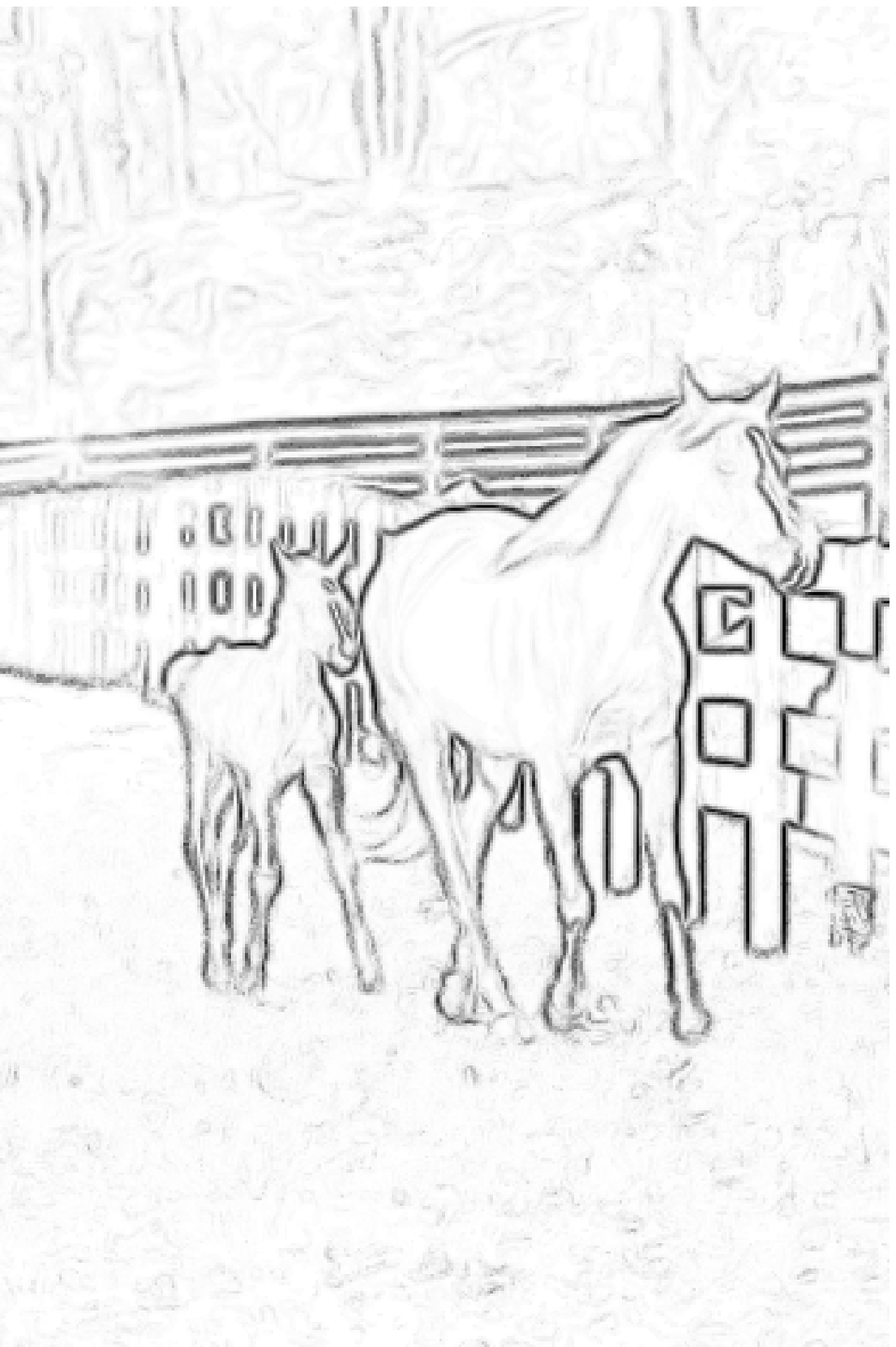} &
\includegraphics[width=0.3\textwidth]{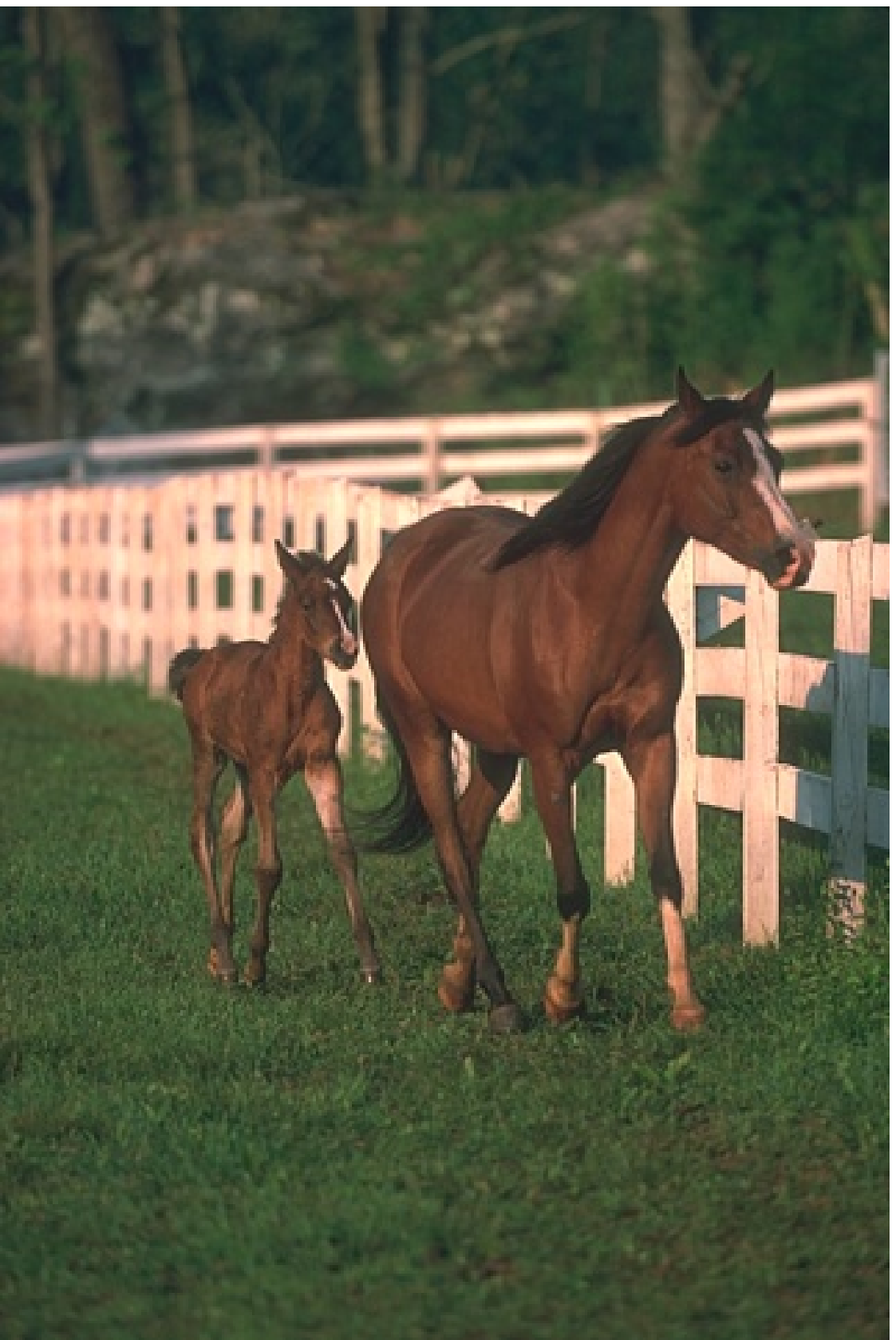} \\
(\coldist) & (IM)
\end{tabular}
\caption{ Edge detection with the generalized compass edge detection \cite{ruzon_compass} using the
  following color differences: (NE) A negative exponent applied on the Euclidean distance in L*a*b*
  space (used in \cite{ruzon_compass}).  (TC) A thresholded CIEDE2000 distance (used in
  \cite{Pele-iccv2009} for image retrieval). See \eqRefText \ref{d1_eq}.  (\coldist) Our
  proposed \coldist.  (IM) The original image. \newline Our results are much cleaner.  }
\label{res_3}
\end{figure*}

\begin{figure*}[htbp] \centering
\begin{tabular}{cc}
\includegraphics[width=0.45\textwidth]{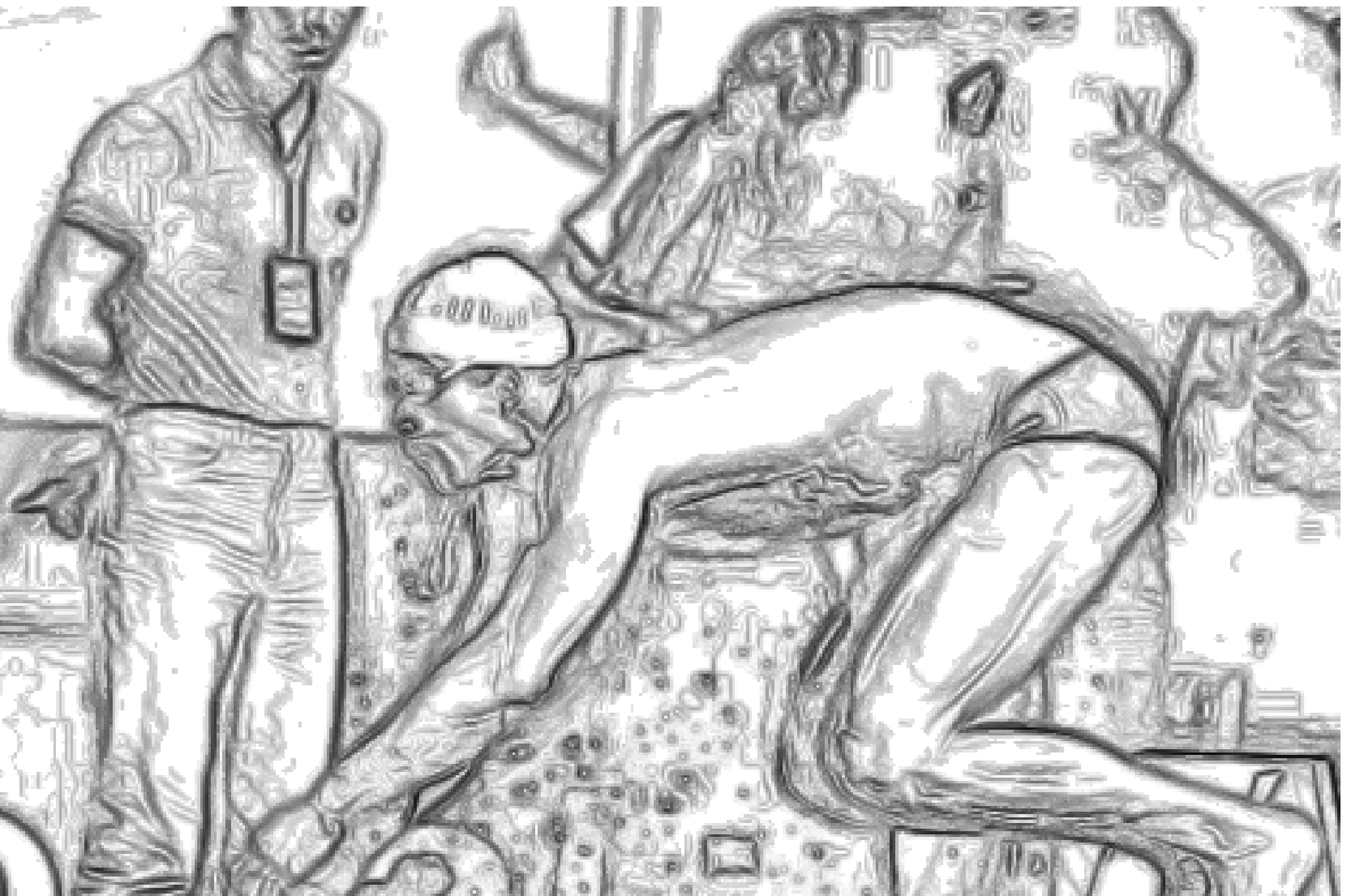} &
\includegraphics[width=0.45\textwidth]{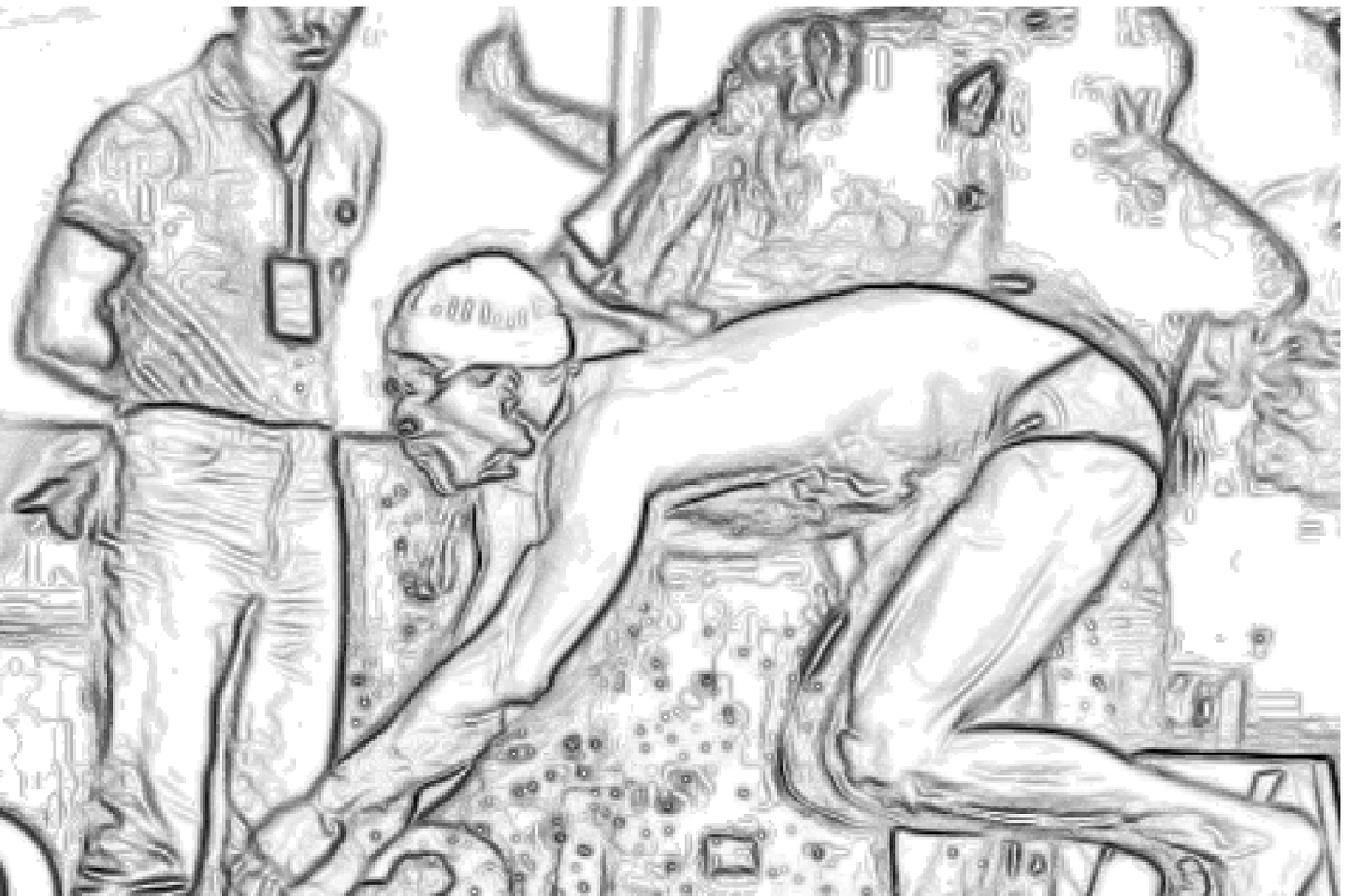} \\
(NE) & (TC) \\
\includegraphics[width=0.45\textwidth]{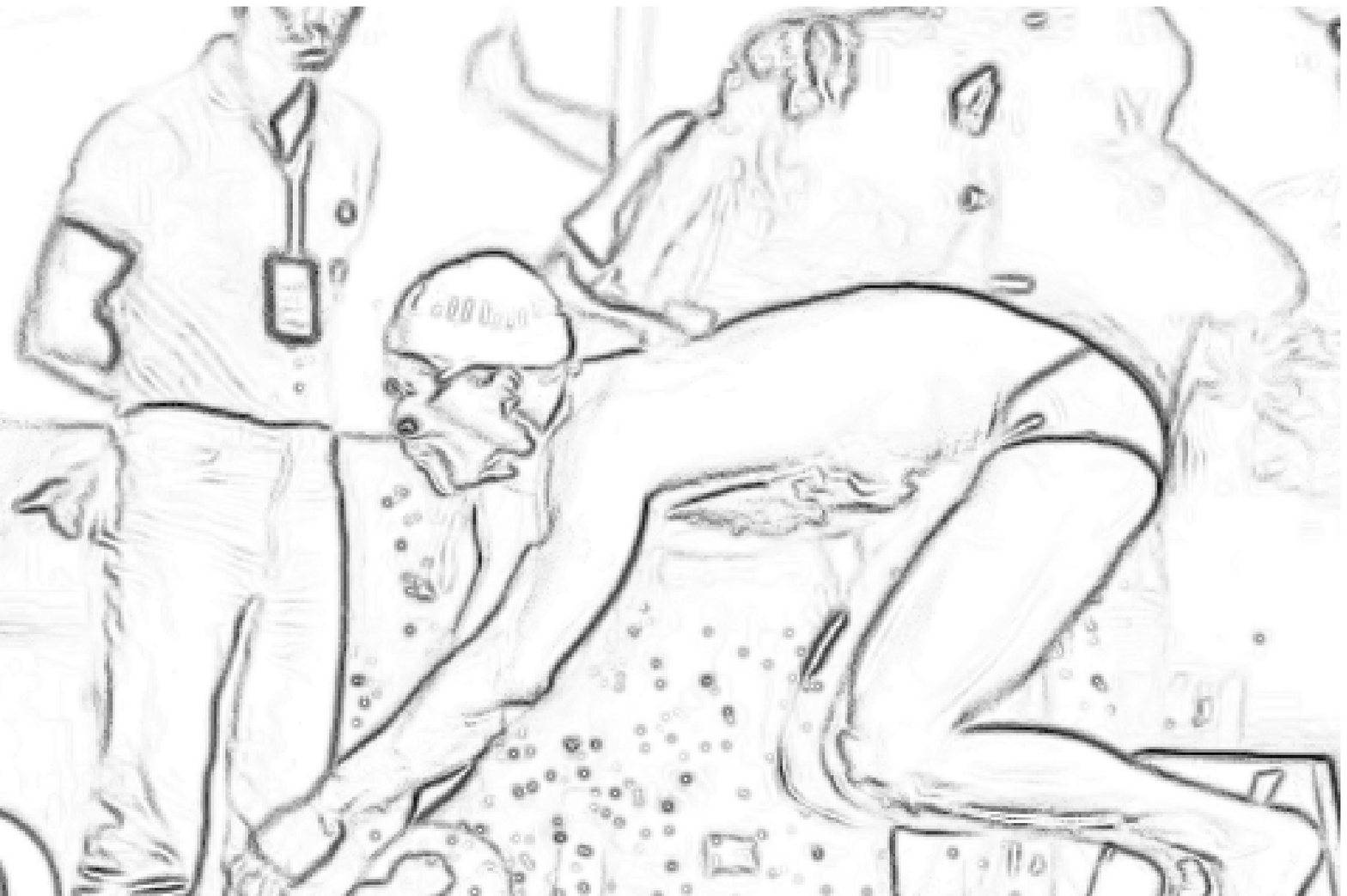} &
\includegraphics[width=0.45\textwidth]{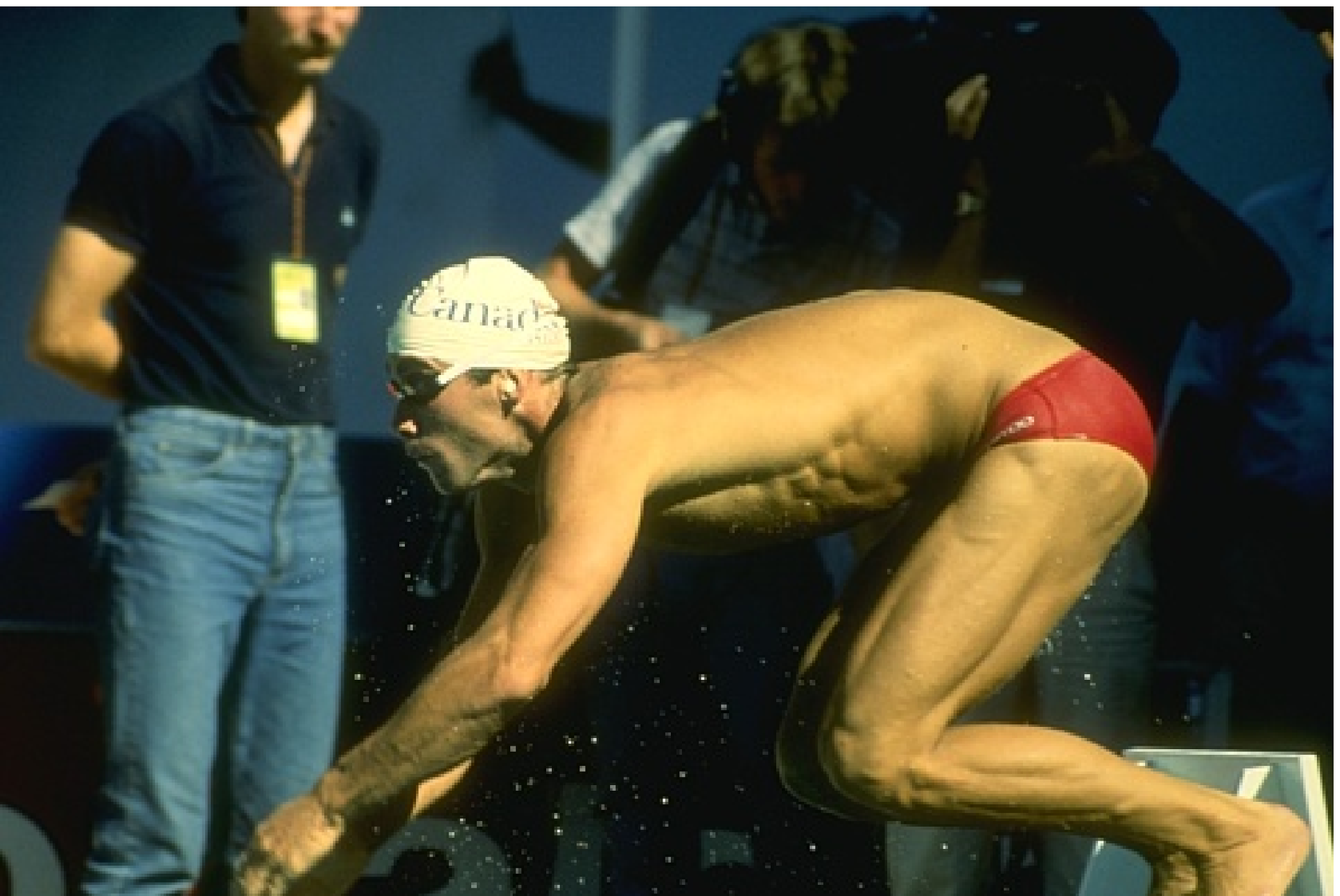} \\
(\coldist) & (IM) \\
\newline \\
\includegraphics[width=0.45\textwidth]{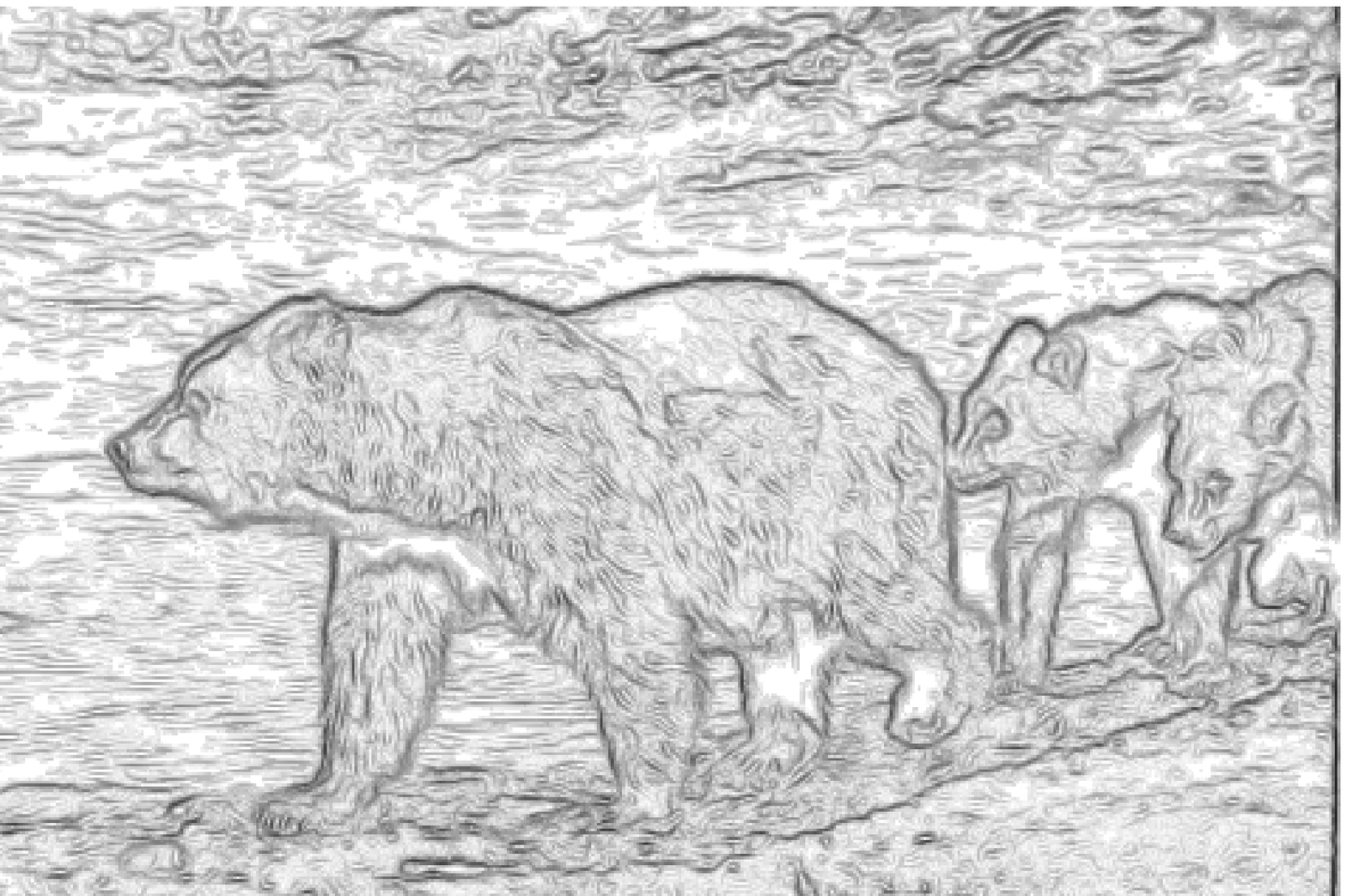} &
\includegraphics[width=0.45\textwidth]{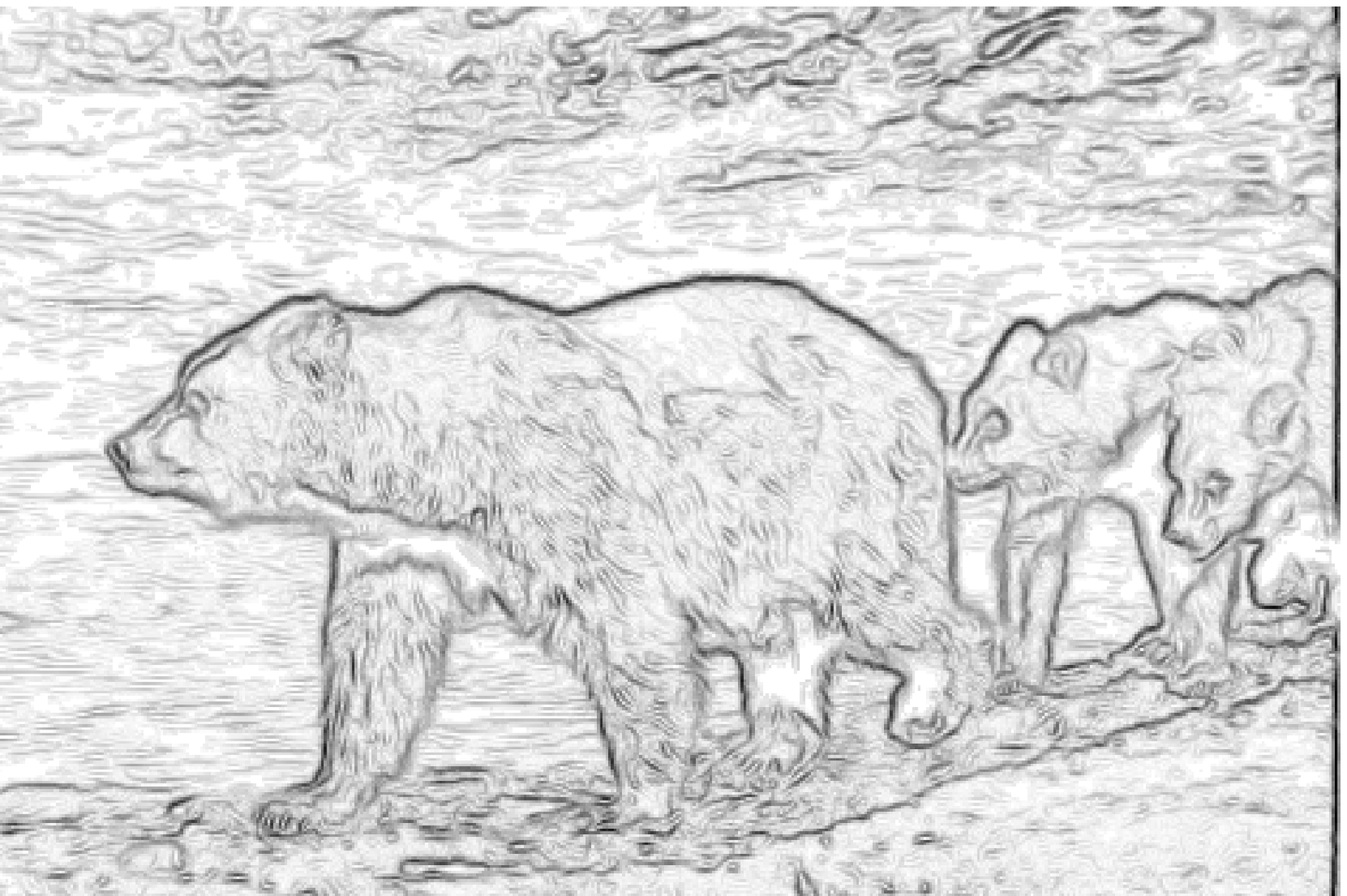} \\
(NE) & (TC) \\
\includegraphics[width=0.45\textwidth]{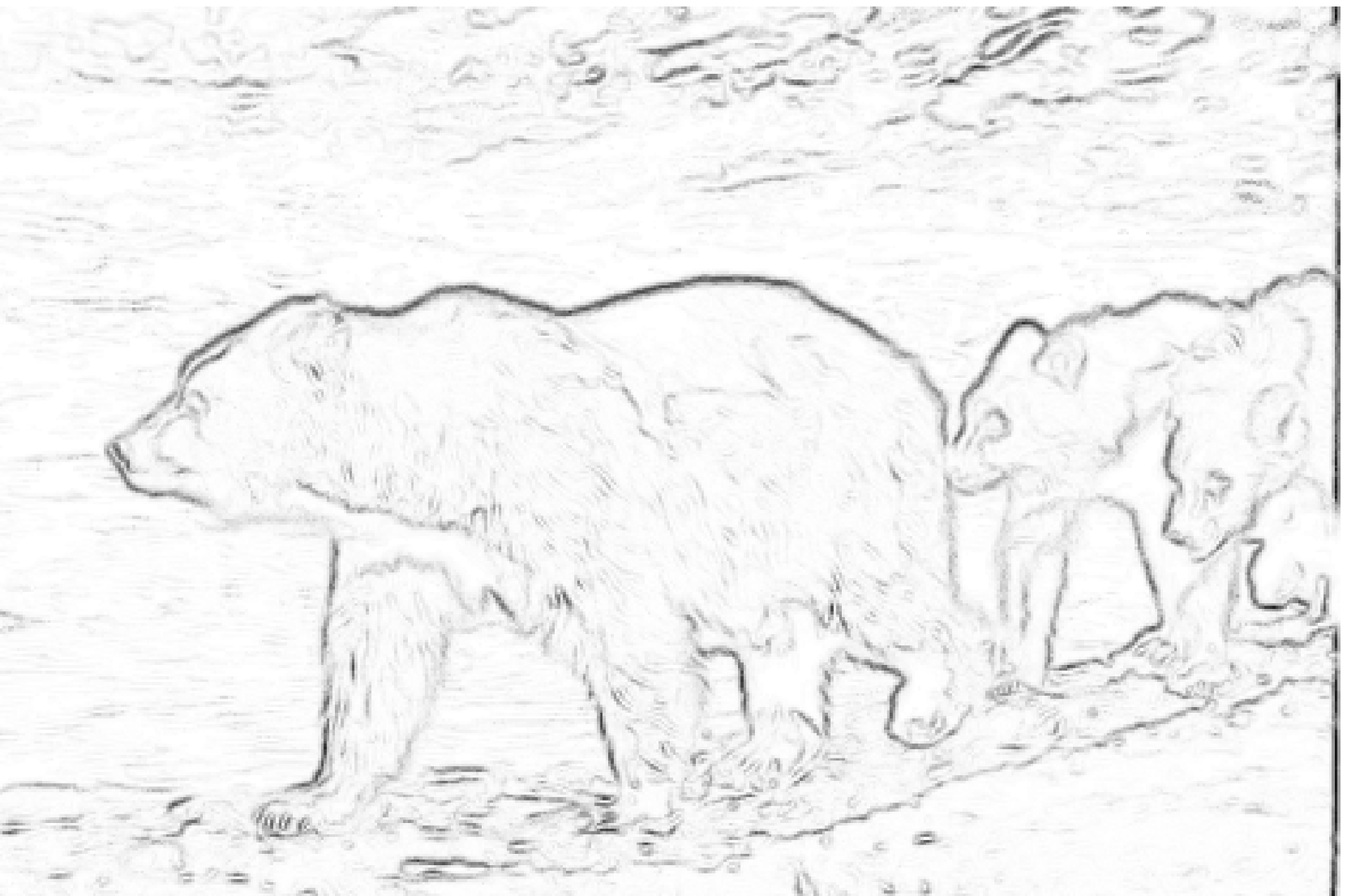} &
\includegraphics[width=0.45\textwidth]{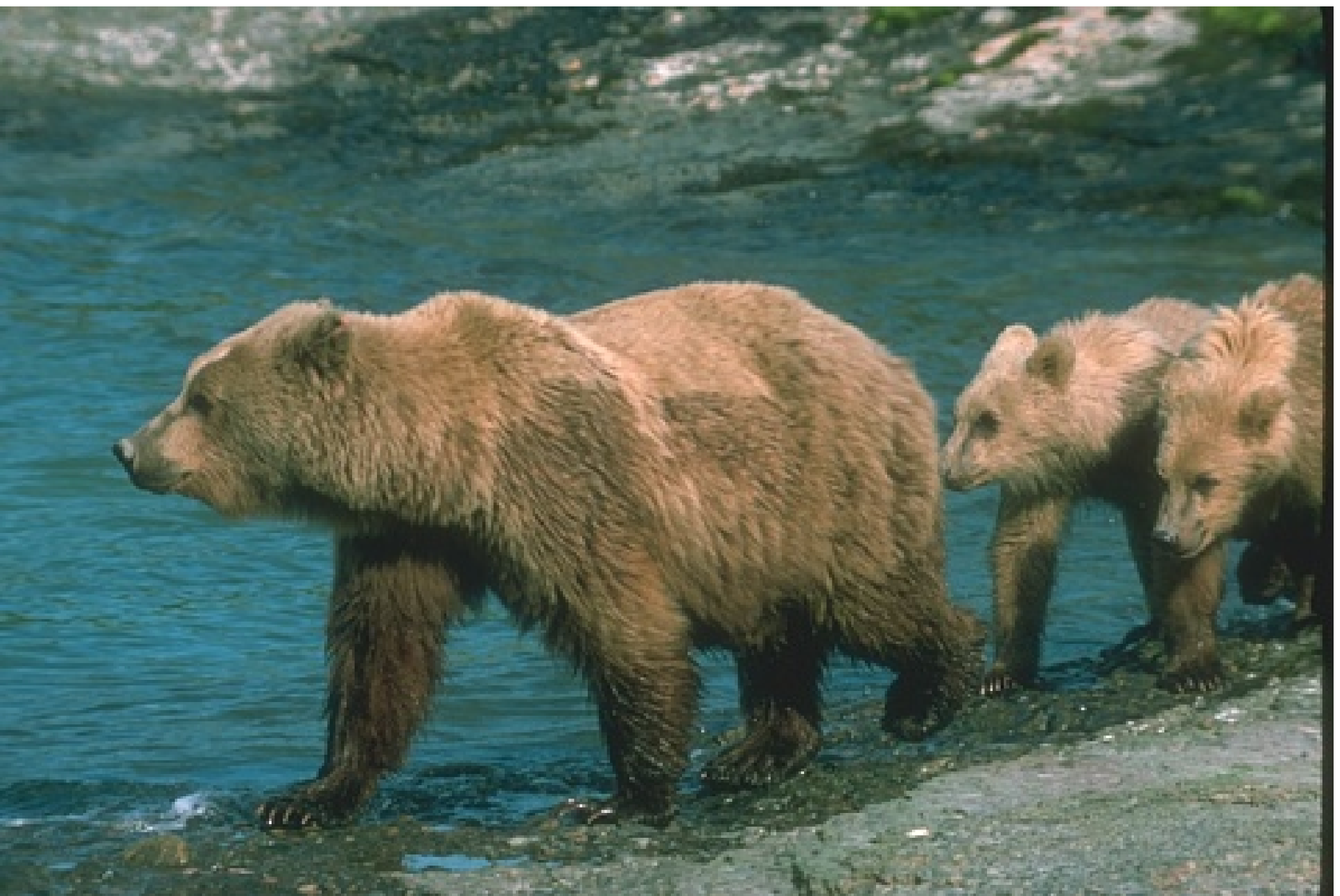} \\
(\coldist) & (IM)
\end{tabular}
\caption{ Edge detection with the generalized compass edge detection \cite{ruzon_compass} using the
  following color differences: (NE) A negative exponent applied on the Euclidean distance in L*a*b*
  space (used in \cite{ruzon_compass}).  (TC) A thresholded CIEDE2000 distance (used in
  \cite{Pele-iccv2009} for image retrieval). See \eqRefText \ref{d1_eq}.  (\coldist) Our proposed
  \coldist.  (IM) The original image. \newline Our results are much cleaner. Note the strong
  responses on the fur of the bear and in the left person on the top T-shirt using (NE) and
  (TC). Note that the spots around the swimmer in all methods are due to successful detection of
  the water drops.}
\label{res_4}
\end{figure*}

Ruzon and Tomasi used $\gamma=14$ in their experiments. We compare the edge detection results using
this distance to our proposed \coldist. We compared also to $d_1$ which is a thresholded CIEDE2000
distance as was used by Pele and Werman for image retrieval \cite{Pele-iccv2009}. We also tried our
proposed \coldist{} without the sigmoid function or without the color correction ($\alpha=1$) or
without the CIEDE2000 term ($\alpha=0$) but the results using \coldist{} were the best.  Results
are presented in \figsRefText \ref{res_1},\ref{res_2},\ref{res_3},\ref{res_4}. The results show that the new color
difference is able to detect color edges much better than the state of the art. The resulting edge
maps are much cleaner. See figures captions for more details.

\section{Conclusions}

We presented a new color difference - \coldist{} and showed that it is perceptually more meaningful
than the state of the art color difference - CIEDE2000. We believe that this is just the first step
in designing perceptual color differences which perform well in the medium range.

It is easy to generalize our method to other color name sets (such as the Russian which separates
blue into goluboi and siniy). All one needs to do is to calculate the ground distance between all
color terms. This can be done by using the joint distribution of the new set of color terms.  In
future work it will be interesting to check other color naming methods such as
\cite{conway1992experimental,lammens1994computational,seaborn1999fuzzy,benaventeBOV00,griffin2004optimality,mojsilovic2005computational,benavente2006data,menegaz2006discrete,menegaz2007semantics,benavente2008parametric}

A major difficulty of analyzing color images is the illumination variability of scenes. Color
invariants are often used to overcome this problem. However, Van de Weijer \etal
\cite{vandeweijer2009lcn,vandeweijer2007acn} showed that invariants are not discriminative
enough. For example, invariants usually do not distinguish between achromatic colors (black, gray and
white). Using color constancy or partial normalization algorithms
\cite{color_constancy_0,color_constancy_1,color_constancy_2,color_constancy_3,color_constancy_4,vdw8,lucolor,bianco2008improving,tancolor}
which do not necessarily reduce all distinctiveness may partially alleviate this problem. This method
was used to improve color naming by Benavente \etal \cite{benaventeBOV00}.

Color perception is also affected by spatial and texture cues. It will be interesting to combine
\coldist{} with spatial and texture models \cite{texture_spatial_color_2,texture_spatial_color_1,texture_spatial_color_3}


\label{conclusions_sec}

{\small
\bibliographystyle{ieee}
\bibliography{color}

\begin{thebibliography}{10}\itemsep=-1pt

\bibitem{wert}
{\em Numbers and numerical concepts in primitive peoples}.

\bibitem{cie_recommendation_5_units}
Cie pub. 142, 2001.

\bibitem{benaventeBOV00}
R.~Benavente, R.~Baldrich, M.~C. Oliv{\'e}, and M.~Vanrell.
\newblock Colour naming considering the colour variability problem.
\newblock {\em Computaci{\'o}n y Sistemas}, 4(1):30--43, 2000.

\bibitem{benavente2006data}
R.~Benavente, M.~Vanrell, and R.~Baldrich.
\newblock {A data set for fuzzy colour naming}.
\newblock {\em COLOR research and application}, 31(1):48, 2006.

\bibitem{benavente2008parametric}
R.~Benavente, M.~Vanrell, and R.~Baldrich.
\newblock {Parametric fuzzy sets for automatic color naming}.
\newblock {\em Journal of the Optical Society of America A}, 25(10):2582--2593,
  2008.

\bibitem{berlin_kay}
B.~Berlin and P.~Kay.
\newblock {\em {Basic Color Terms: Their Universality and Evolution}}.
\newblock University of California Press Berkeley, 1969.

\bibitem{bianco2008improving}
S.~Bianco, G.~Ciocca, C.~Cusano, and R.~Schettini.
\newblock {Improving Color Constancy Using Indoor--Outdoor Image
  Classification}.
\newblock {\em IEEE Transactions on Image Processing}, 17(12):2381--2392, 2008.

\bibitem{BurghoutsCVIU2009}
G.~J. Burghouts and J.~M. Geusebroek.
\newblock Performance evaluation of local colour invariants.
\newblock {\em Computer Vision and Image Understanding}, 113:48--62, 2009.

\bibitem{conway1992experimental}
D.~Conway.
\newblock {An experimental comparison of three natural language colour naming
  models}.
\newblock In {\em Proc. east-west int. conf. on human-computer interaction},
  pages 328--339. Citeseer, 1992.

\bibitem{vdw8}
G.~Finlayson, B.~Schiele, and J.~Crowley.
\newblock {Comprehensive colour image normalization}.
\newblock In {\em ECCV}, volume 1406, page 1406, 1998.

\bibitem{color_constancy_4}
G.~D. Finlayson, S.~D. Hordley, and P.~M. Morovic.
\newblock Colour constancy using the chromagenic constraint.
\newblock In {\em CVPR}, pages 1079--1086, 2005.

\bibitem{color_constancy_0}
D.~Forsyth.
\newblock {A novel algorithm for color constancy}.
\newblock In {\em Color}, page 271. Jones and Bartlett Publishers, Inc., 1992.

\bibitem{color_constancy_1}
P.~V. Gehler, C.~Rother, A.~Blake, T.~Minka, and T.~Sharp.
\newblock Bayesian color constancy revisited.
\newblock 2008.

\bibitem{vdw12}
T.~Gevers and H.~Stokman.
\newblock {Robust histogram construction from color invariants for object
  recognition}.
\newblock {\em IEEE transactions on pattern analysis and machine intelligence},
  26(1):113--118, 2004.

\bibitem{color_constancy_2}
A.~Gijsenij and T.~Gevers.
\newblock Color constancy using natural image statistics.
\newblock In {\em CVPR}, 2007.

\bibitem{griffin2004optimality}
L.~Griffin.
\newblock {Optimality of the basic colours categories}.
\newblock {\em Journal of Vision}, 4(8):309, 2004.

\bibitem{large_color_diffs}
M.~R.~L. Han~Wang, Guihua~Cui and H.~Xu.
\newblock Evaluation of colour-difference formulae for different
  colour-difference magnitudes.
\newblock In {\em CGIV}, 2008.

\bibitem{indow1994metrics}
T.~Indow.
\newblock {Metrics in Color Spaces: Im Kleinen und im Grofien}.
\newblock {\em Contributions to mathematical psychology, psychometrics, and
  methodology}, page~3, 1994.

\bibitem{lammens1994computational}
J.~Lammens.
\newblock {\em {A computational model of color perception and color naming}}.
\newblock PhD thesis, Citeseer, 1994.

\bibitem{lucolor}
R.~Lu, A.~Gijsenij, T.~Gevers, D.~Xu, V.~Nedovic, and J.~Geusebroek.
\newblock {Color Constancy Using 3D Stage Geometry}.
\newblock In {\em IEEE International Conference on Computer Vision}.

\bibitem{ciede2000}
M.~Luo, G.~Cui, and B.~Rigg.
\newblock {The Development of the CIE 2000 Colour-Difference Formula:
  CIEDE2000}.
\newblock {\em Color Research \& Application}, 26(5):340--350, 2001.

\bibitem{macadam1942vsc}
D.~MacAdam.
\newblock {Visual sensitivities to color differences in daylight}.
\newblock {\em Journal of the Optical Society of America}, 32(5):247--273,
  1942.

\bibitem{menegaz2007semantics}
G.~Menegaz, A.~Le~Troter, J.~Boi, and J.~Sequeira.
\newblock {Semantics driven resampling of the osa-ucs}.
\newblock In {\em Image Analysis and Processing Workshops, 2007. ICIAPW 2007.
  14th International Conference on}, pages 216--220, 2007.

\bibitem{menegaz2006discrete}
G.~Menegaz, A.~Le~Troter, J.~Sequeira, and J.~Boi.
\newblock {A discrete model for color naming}.
\newblock {\em EURASIP Journal on Advances in Signal Processing}, 2007, 2006.

\bibitem{vdw13}
F.~Mindru, T.~Tuytelaars, L.~Gool, and T.~Moons.
\newblock {Moment invariants for recognition under changing viewpoint and
  illumination}.
\newblock {\em Computer Vision and Image Understanding}, 94(1-3):3--27, 2004.

\bibitem{mojsilovic2005computational}
A.~Mojsilovic.
\newblock {A computational model for color naming and describing color
  composition of images}.
\newblock {\em IEEE Transactions on Image Processing}, 14(5):690--699, 2005.

\bibitem{Pele-iccv2009}
O.~Pele and M.~Werman.
\newblock Fast and robust earth mover's distances.
\newblock In {\em ICCV}, 2009.

\bibitem{peleeccv2010}
O.~Pele and M.~Werman.
\newblock The quadratic-chi histogram distance family.
\newblock In {\em ECCV}, 2010.

\bibitem{robertson1990hdc}
A.~Robertson.
\newblock {Historical development of CIE recommended color difference
  equations}.
\newblock {\em Color Res. Appl}, 15:167--170, 1990.

\bibitem{rosch1975cognitive}
E.~Rosch.
\newblock {Cognitive reference points* 1}.
\newblock {\em Cognitive psychology}, 7(4):532--547, 1975.

\bibitem{rubner_emd_comparison}
Y.~Rubner, J.~Puzicha, C.~Tomasi, and J.~Buhmann.
\newblock {Empirical evaluation of dissimilarity measures for color and
  texture}.
\newblock {\em CVIU}, 2001.

\bibitem{rubner_emd}
Y.~Rubner, C.~Tomasi, and L.~J. Guibas.
\newblock The earth mover's distance as a metric for image retrieval.
\newblock {\em International Journal of Computer Vision}, 40(2):99--121, 2000.

\bibitem{ruzon_compass}
M.~Ruzon and C.~Tomasi.
\newblock {Edge, Junction, and Corner Detection Using Color Distributions}.
\newblock {\em IEEE Trans. Pattern Analysis and Machine Intelligence.}, pages
  1281--1295, 2001.

\bibitem{seaborn1999fuzzy}
M.~Seaborn, L.~Hepplewhite, and J.~Stonham.
\newblock {Fuzzy colour category map for content based image retrieval}.
\newblock In {\em BMVC}, 1999.

\bibitem{sharma2005ccd}
G.~Sharma, W.~Wu, and E.~Dalal.
\newblock {The CIEDE2000 color-difference formula: implementation notes,
  supplementary test data, and mathematical observations}.
\newblock {\em Color Research \& Application}, 30(1):21--30, 2005.

\bibitem{songlocal}
X.~Song, D.~Muselet, and A.~Tr{\'e}meau.
\newblock {Local Color Descriptor for Object Recognition across Illumination
  Changes}.
\newblock In {\em Advanced Concepts for Intelligent Vision Systems}, pages
  598--605. Springer.

\bibitem{tancolor}
R.~Tan, K.~Ikeuchi, and K.~Nishino.
\newblock {Color constancy through inverse-intensity chromaticity space}.
\newblock {\em Digitally Archiving Cultural Objects}, pages 323--351.

\bibitem{texture_spatial_color_3}
O.~Tulet, M.-C. Larabi, and C.~Fernandez-Maloigne.
\newblock Image rendering based on a spatial extension of the ciecam02.
\newblock {\em Applications of Computer Vision, IEEE Workshop on}, 2008.

\bibitem{color_features_1}
K.~E.~A. van~de Sande, T.~Gevers, and C.~G.~M. Snoek.
\newblock Evaluation of color descriptors for object and scene recognition.
\newblock In {\em CVPR}, 2008.

\bibitem{vdw11}
J.~van~de Weijer, T.~Gevers, and J.~Geusebroek.
\newblock {Edge and corner detection by photometric quasi-invariants}.
\newblock {\em IEEE transactions on pattern analysis and machine intelligence},
  27(4):625--630, 2005.

\bibitem{color_constancy_3}
J.~van~de Weijer, T.~Gevers, and A.~Gijsenij.
\newblock Edge-based color constancy.
\newblock {\em IEEE Transactions on Image Processing}, 16(9):2207--2214, 2007.

\bibitem{vandeweijer2006clf}
J.~van~de Weijer and C.~Schmid.
\newblock {Coloring local feature extraction}.
\newblock In {\em ECCV}, page 334, 2006.

\bibitem{vandeweijer2007acn}
J.~van~de Weijer and C.~Schmid.
\newblock {Applying Color Names to Image Description}.
\newblock In {\em ICIP}, volume~3, 2007.

\bibitem{vandeweijer2009lcn}
J.~van~de Weijer, C.~Schmid, J.~Verbeek, and D.~Larlus.
\newblock {Learning Color Names for Real-World Applications}.
\newblock {\em IEEE Transaction in Image Processing}, 2009.

\bibitem{texture_spatial_color_1}
M.~Vanrell, R.~Baldrich, A.~Salvatella, R.~Benavente, and F.~Tous.
\newblock Induction operators for a computational colour-texture
  representation.
\newblock {\em Computer Vision and Image Understanding}, 94(1-3):92--114, 2004.

\bibitem{color_book}
G.~Wyszecki and W.~S. Stiles.
\newblock {\em Color Science: Concepts and Methods, Quantitative Data and
  Formulae}.
\newblock Wiley, 1982.

\bibitem{ciede2000_test_on_crt}
H.~Xu, H.~Yaguchi, and S.~Shioiri.
\newblock {Estimation of Color-Difference Formulae at Color Discrimination
  Threshold Using CRT-Generated Stimuli}.
\newblock {\em Optical Review}, 8(2):142--147, 2001.

\bibitem{texture_spatial_color_2}
X.~Zhang and B.~Wandell.
\newblock {A spatial extension of CIELAB for digital color-image reproduction}.
\newblock {\em Journal of the Society for Information Display}, 5:61, 1997.

\end{thebibliography}
}

\end{document}